\documentclass[letterpaper]{article}
\usepackage[preprint]{aaai2027}
\usepackage[hyphens]{url} 
\usepackage{graphicx} 
\usepackage{natbib} 
\usepackage{caption} 
\usepackage{amsmath}
\usepackage{amsfonts}
\usepackage{amssymb}
\usepackage{bm}
\usepackage{nicefrac}

\usepackage{algorithm}
\usepackage{algorithmic}

\usepackage{booktabs}
\usepackage{multirow}
\usepackage{subcaption}
\usepackage{xcolor}
\usepackage{colortbl}
\usepackage{float}
\usepackage{array}
\usepackage{enumitem}
\usepackage{needspace}
\definecolor{graybg}{gray}{0.95}

\graphicspath{{figure/}}

\newcommand{\wrstd}[2]{#1{\scriptsize$\pm$#2}}
\newcommand{\best}[1]{\textbf{#1}}
\newcommand{\second}[1]{\underline{#1}}
\newcommand{\nop}[1]{}
\newcommand{\method}{our method}
\newcommand{\ind}{\mathbb{I}}
\newcommand{\diag}{\operatorname{Diag}}

\title{Learning to Adapt Cross-Domain Preferences via Meta-LoRA for LLM Personalization}

\author{
\textbf{Xuefei Wang}\textsuperscript{\rm 1},
\textbf{Jun Han}\textsuperscript{\rm 1},
\textbf{Zixuan Wang}\textsuperscript{\rm 1},
\textbf{Qingkai Zeng}\textsuperscript{\rm 2},\\
\textbf{Xiao Wang}\textsuperscript{\rm 1},
\textbf{Ruijie Wang}\textsuperscript{\rm 1}\corresponding,
\textbf{Jianxin Li}\textsuperscript{\rm 1}
}
\affiliations{
\textsuperscript{\rm 1}Beihang University \qquad
\textsuperscript{\rm 2}Nankai University\\
\texttt{xuefeiw@buaa.edu.cn}, \texttt{ruijiew@buaa.edu.cn}
}

\begin{document}

\maketitle

\begin{abstract}
Cross-domain zero- or few-shot personalization aims to generate user-preferred responses in unseen conversational domains from only a handful of target-domain interactions. 
Existing adaptation methods struggle to calibrate update magnitude under sparse evidence and thus overfit, whereas history-transfer methods often entangle user preferences with source-domain artifacts, yielding unreliable personalization priors and negative transfer. 
To calibrate adaptation to evidence quality, we propose PAC-Bayes-regularized Meta-LoRA, which uses a meta-learned LoRA initialization as both the adaptation start and prior center, while adjusting update strength according to support-set size and predictive uncertainty. This limits overfitting under sparse or ambiguous evidence while permitting stronger personalization as evidence grows.
Controlled adaptation alone does not determine which preferences should transfer across domains or how they should be expressed. We therefore functionally decompose personalization priors into user and domain components, using a human-readable prompt for stable preferences and topology-preserving soft tokens for domain-specific hidden-space conditioning.
Experiments across multiple benchmarks and personalization tasks show consistent gains over strong baselines. On HiCUPID, our method reduces cross-domain win-rate degradation by 47.9\% relative to the best competing baseline and improves win rate by 110.2\% under unseen-user cold start.

\end{abstract}

\section{Introduction}\label{sec:intro}

Large Language Models (LLMs) have demonstrated strong general capabilities across a wide range of tasks. However, a persistent gap remains between their generic competence and user satisfaction at the individual level \citep{salemi2025lamp,mok2025exploring}. 
This gap arises from the inherently personalized nature of human interaction: user preferences are high-dimensional, latent, and context-dependent \citep{gao-etal-2026-beyond}. 
They shape style, reasoning depth, evidence use, and constraint handling. 
Personalized response generation therefore requires more than retrieving user facts or imitating past wording \citep{liu-etal-2025-llms}: it must identify stable user tendencies and realize them under the conventions of the current conversational domain. Reliably transferring such preferences across domains remains a central challenge for user-level personalization~\citep{ling2025domain}.

\begin{figure}[t]
    \centering
    \includegraphics[width=0.85\columnwidth]{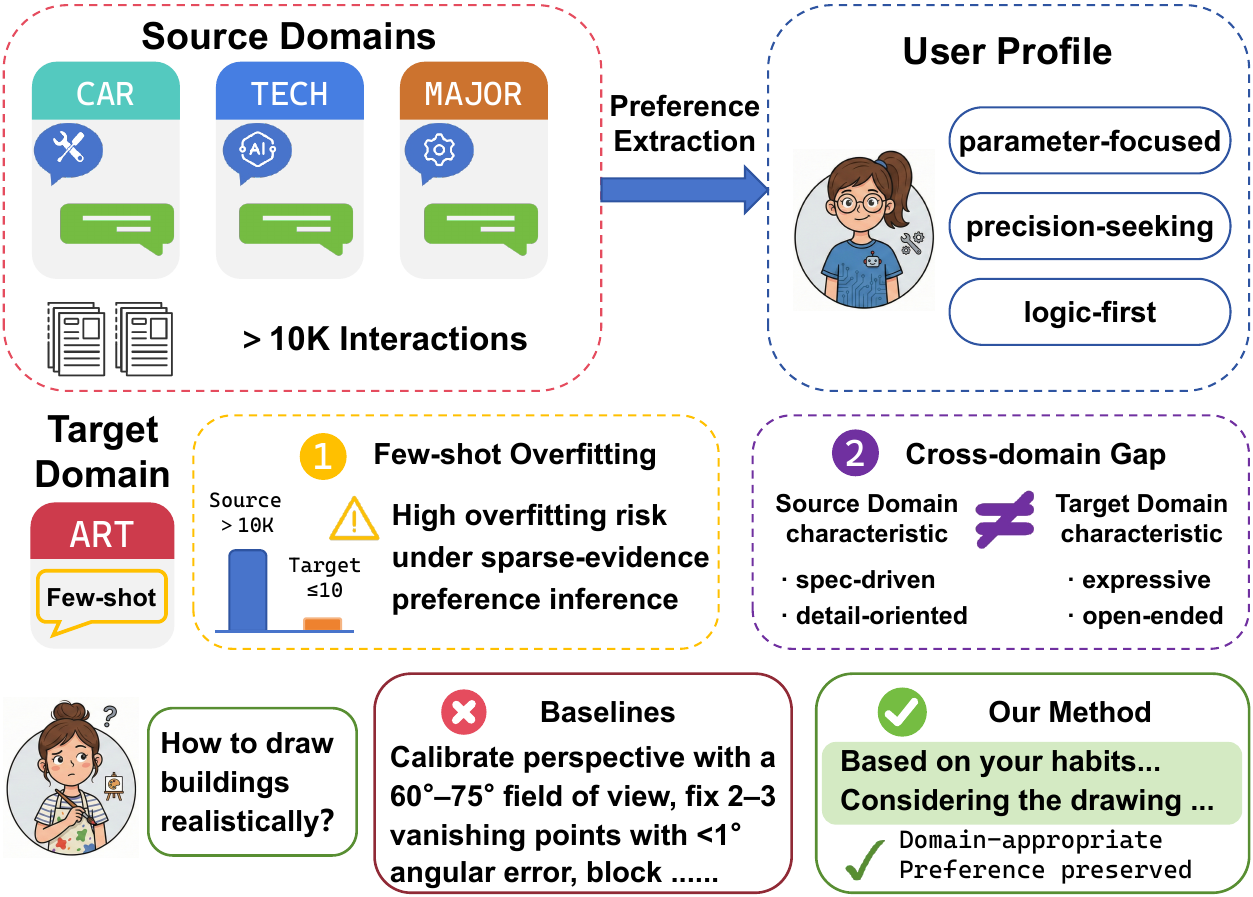}
    \caption{Illustration of the cross-domain LLM personalization problem, which presents few-shot overfitting and cross-domain gap challenges.}
    \label{fig:cross_domain}
    \vspace{-5mm}
\end{figure}

This challenge is particularly acute when target-domain interactions are scarce or absent~\citep{tan-etal-2024-democratizing}. Under zero- or few-shot conditions, limited observations provide incomplete and potentially ambiguous evidence about a user's target-domain preferences, making adaptation vulnerable to overfitting and preference distortion~\citep{ling2025domain,zhang2023collaborative}. Existing methods lack principled control over update magnitude: aggressive updates may overwrite transferable personalization patterns, whereas conservative updates fail to capture user-specific patterns~\citep{choi-etal-2025-copl}. Cross-domain heterogeneity further complicates transfer~\citep{gao-etal-2026-beyond}. The same preference may require different realizations across domains with distinct discourse conventions and pragmatic expectations; consequently, history-based representations may entangle stable user tendencies with source-domain artifacts, causing negative transfer or preference drift. Together, evidence scarcity and domain heterogeneity remain key obstacles to robust cross-domain personalization, as illustrated in Figure~\ref{fig:cross_domain}.

Prior work personalizes LLMs through retrieval and profiling, parameter adaptation, personalized alignment, and inference-time steering. Retrieval- and profile-based methods may transfer source-specific content rather than persistent preferences~\citep{salemi2024lamp,zhang2024guided,wang2024crafting,nam2025learning}. Parameter-efficient fine-tuning and meta-learning can be dominated by atypical sparse examples and miscalibrate adaptation under domain shift~\citep{choi2025copl,zhang2025proper,singh2025fspo,tan2025instant}. Alignment and steering control user-level behavior but rarely model its realization under unseen-domain conventions~\citep{cao2024personalized,chen2024pad,kim2025drift,zhu2025fly}. These paradigms therefore leave two related problems insufficiently addressed: adapting parameters according to scarce target evidence and constructing personalization priors that transfer reliably across domains.

To address these problems, we propose a unified framework that combines PAC-Bayes-regularized Meta-LoRA with dual-channel prior conditioning. Episodic meta-learning over source domain tasks yields a LoRA initialization that captures transferable adaptation structure. We use this initialization both as the starting point for adaptation and as the center of the PAC-Bayes prior, thereby encouraging target-domain updates to preserve transferable behavior and deviate only when supported by user evidence. The anchoring strength is calibrated by adaptation evidence quality. Throughout meta-training and target adaptation, the model is conditioned on personalization priors, allowing limited target interactions to refine the personalization context rather than infer user preferences from scratch.

Constructing these priors requires distinguishing what should transfer across domains from how it should be expressed in the target domain. We therefore functionally decompose personalization priors into user and domain components and match each to an appropriate conditioning interface. Stable user preferences are injected through a human-readable prompt, preserving interpretability, while target-domain conditions are injected as compact soft tokens that provide fine-grained hidden-space control. This dual-channel design avoids appending long histories and allows the same user preference to be realized differently across domains.

Our main contributions are threefold:
\begin{itemize}
    \item We introduce PAC-Bayes-regularized Meta-LoRA, which uses a meta-learned LoRA initialization as both adaptation start and prior center, calibrating update strength by support size and predictive uncertainty.

    \item We propose functional user-domain prior factorization with dual-channel conditioning, injecting readable user preferences through prompts and compact domain conditions through soft tokens.

    \item Experiments show consistent improvements across cross-domain, few-shot, distant-domain, and unseen-user settings, reducing cross-domain win-rate degradation by 47.9\% and improving unseen-user win rate by 110.2\%.
\end{itemize}

\begin{figure*}[t]
    \centering
    \includegraphics[width=\textwidth]{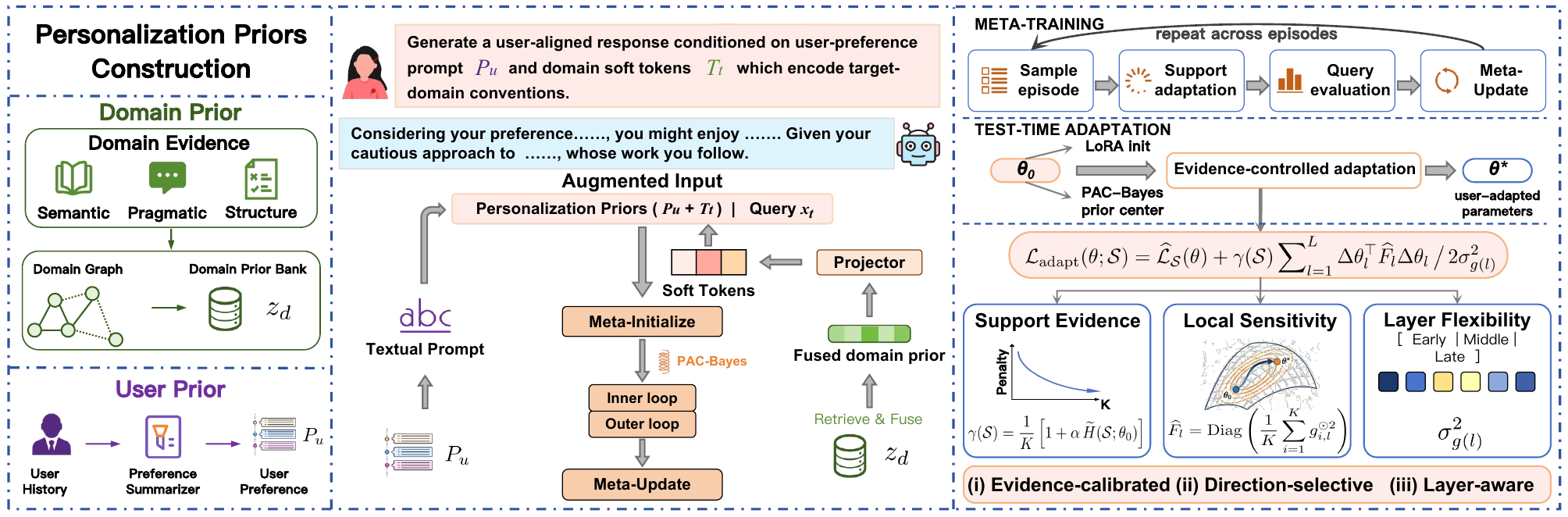}
    \setlength{\abovecaptionskip}{2pt}
    \vspace{-2mm}
    \caption{Framework overview. User history yields a textual prompt $P_u$, while graph-guided domain composition produces soft tokens $T_t$; both augment the input, while PAC-Bayes-regularized Meta-LoRA controls target-domain parameter adaptation.}
    \label{fig:framework_overview}
    \vspace{-3mm}
\end{figure*}
\section{Related Work}

\paragraph{LLM Personalization.}
Existing methods condition LLMs on retrieved interactions, textual profiles, continuous prompts, user-specific parameters, or inference-time preference directions~\citep{salemi-etal-2024-lamp,liu-etal-2025-llms,lester-etal-2021-power,li-liang-2021-prefix,tan-etal-2024-democratizing,NEURIPS2024_58cbe393}. Retrieval and profiles are transparent but may carry source-specific topics or wording into a new domain, while continuous prompts and adapters provide expressive conditioning but may entangle recurring user tendencies with their observed domain-specific realizations~\citep{qiu-etal-2025-latent,choi-etal-2025-copl}. Alignment and decoding-time steering control personalized behavior without parameter updates~\citep{balepur-etal-2025-whose,bu-etal-2025-personalized}, but primarily target observed preferences rather than their realization under unseen-domain conventions. Our work instead assigns user tendencies and domain realization conditions distinct functional roles and conditioning interfaces.

\paragraph{Meta-Learning and Controlled Adaptation.}
Meta-learning learns transferable initializations or update rules for rapid adaptation from limited task data~\citep{pmlr-v70-finn17a}. Personalized parameter-efficient tuning similarly adapts compact user-specific parameters while freezing the backbone~\citep{hu2022lora,tan-etal-2024-democratizing}. These approaches emphasize adaptation speed and support-set fit, but typically do not regulate parameter displacement according to support-set size and model uncertainty. PAC-Bayes analysis relates empirical risk to prior--posterior complexity~\citep{mcallester1999some}, providing a principled basis for such control. We instantiate this connection by centering the parameter prior at a source-learned LoRA initialization and using the resulting complexity structure to guide evidence-calibrated target adaptation.
\section{Methodology}

Given a user $u$ with source-domain history $\mathcal{H}_u$ observed during training, we consider a target domain $d_t$ that is unseen during training. Our goal is to generate a user-aligned response $y_t$ to a query $x_t$ in the sparsely observed target domain $d_t$. Let $\mathcal{S}=\{(x_i,y_i)\}_{i=1}^{K}$ denote the target-domain support set, where $K=0$ and $K>0$ correspond to the zero-shot and few-shot settings, respectively. Complete task and information-access definitions are in Appendix~B.

We propose a unified framework consisting of PAC-Bayes-Regularized Meta-LoRA and dual-channel personalization priors, as illustrated in Figure~\ref{fig:framework_overview}.
Episodic meta-learning first learns a transferable LoRA initialization $\theta_0$, used directly with the injected priors in the zero-shot setting and as the starting point for a few inner-loop updates on $\mathcal{S}$ in the few-shot setting. We further use $\theta_0$ as the center of a PAC-Bayes parameter prior and calibrate the adaptation strength to the available support evidence, limiting excessive parameter drift under sparse or uncertain observations. 

The personalization condition is jointly constructed through two functionally complementary channels: a user-side prior and a domain-side prior. The former is derived from $\mathcal{H}_u$ and injected as a textual prompt, whereas the latter is constructed from source-domain characteristics and target-context evidence and injected as soft tokens. Together, they enable user preferences to be expressed appropriately under the conventions of the target domain.

\subsection{PAC-Bayes-Regularized Meta-LoRA}

When target-domain support examples are scarce or ambiguous, direct fine-tuning may overfit incidental patterns and impair its generalization to future interactions. Although standard Meta-LoRA provides an initialization for rapid adaptation, its inner loop typically optimizes only the support loss and does not regulate how far parameters should move under limited evidence. We therefore anchor target domain LoRA updates to the meta-initialization and adapt the strength and geometry of this anchor to the available evidence.

\paragraph{Meta-Learned Prior and PAC-Bayes-Guided Constraint.}
Let $\theta=\{\theta_l\}_{l=1}^{L}$ denote the LoRA parameters inserted into $L$ Transformer layers, let $\theta_0=\{\theta_{0,l}\}_{l=1}^{L}$ be the initialization learned through source-domain episodic meta-training, and let $g(l)$ index the depth group containing layer $l$. Our goal is to retain the rapid adaptation enabled by $\theta_0$ while preventing a small target support set from driving the model toward incidental patterns. PAC-Bayes is well suited to this objective because it jointly relates support-set fit, deviation from a target-support-independent prior, and the number of available examples. We therefore use $\theta_0$ as both the starting point for task-specific adaptation and the center of a Gaussian parameter prior:
\begin{equation}
\begin{array}{c}
\displaystyle
P_0
=
\prod_{l=1}^{L}
\mathcal{N}\!\bigl(\theta_{0,l},\,\sigma_{g(l)}^2 I\bigr)
\;\;
Q_\theta
=
\prod_{l=1}^{L}
\mathcal{N}\!\bigl(\theta_l,\,\sigma_{g(l)}^2 I\bigr),
\\[4pt]
\displaystyle
\mathrm{KL}\!\bigl(Q_\theta\Vert P_0\bigr)
=
\sum_{l=1}^{L}
\frac{\lVert\theta_l-\theta_{0,l}\rVert_2^2}
     {2\,\sigma_{g(l)}^2}.
\end{array}
\label{eq:pb_prior}
\end{equation}
The depth grouping, $\theta_0$, and the group-wise variances $\{\sigma_g^2\}$ are learned exclusively from source-domain tasks and fixed before $\mathcal{S}$ is observed. Conditioned on the source training data, $P_0$ is therefore independent of the target support set, and its KL complexity measures a depth-group-weighted displacement from the transferable meta-initialization.

For the PAC-Bayes analysis, we map the token-averaged negative log-likelihood to $[0,1]$ using a monotone mapping fixed from source-domain data; actual adaptation continues to optimize the original negative log-likelihood. Consider a support set $\mathcal S=\{(x_i,y_i)\}_{i=1}^{K}$ of $K\geq1$ interactions drawn independently from the target user--domain distribution. Let $\widehat R_{\mathcal S_t}(Q_\theta)$ denote the empirical risk of the randomized predictor obtained by sampling parameters from $Q_\theta$, and let $R(Q_\theta)$ denote its expected risk on a future interaction. For any $\delta\in(0,1)$, with probability at least $1-\delta$ over $\mathcal{S}$, the following standard PAC-Bayes bound holds simultaneously for all $Q_\theta$ in the above family \cite{maurer2004note}:
\begin{equation}
R(Q_\theta)
\leq
\hat R_{\mathcal S}(Q_\theta)
+
\sqrt{
\frac{
\mathrm{KL}(Q_\theta\Vert P_0)
+
\ln\!\bigl((K{+}1)/\delta\bigr)
}{
2K
}
}.
\label{eq:pb_bound}
\end{equation}
Thus, the same displacement from $\theta_0$ contributes more strongly to the bound when fewer support examples are available. Let $A_\theta=\operatorname{KL}(Q_\theta\Vert P_0)+\ln((K+1)/\delta)$. For any $\beta>0$, $\sqrt{A_\theta/(2K)}\leq A_\theta/(4\beta K)+\beta/2$. For fixed $K$, $\delta$, and $\beta$, removing terms independent of $\theta$ leaves $\operatorname{KL}(Q_\theta\Vert P_0)/(4\beta K)$, motivating the $K^{-1}$-scaled anchor in the adaptation objective below. Since response generation uses the posterior mean $\theta$ rather than samples from $Q_\theta$, this derivation determines the prior-centered geometry and support-size dependence of our deterministic adaptation surrogate.

\paragraph{Evidence- and Structure-Aware Adaptation Objective.}
The preceding PAC-Bayes analysis relates adaptation complexity to the amount of target evidence, but support-set size alone does not indicate how informative that evidence is for the current model. We therefore characterize effective adaptation evidence through two complementary signals: $K$ provides a coarse measure of preference-evidence coverage, while predictive entropy captures model-relative uncertainty about the observed responses. Consistent with uncertainty-aware calibration, we treat entropy as a property of the model's predictive distribution \citep{krishnan2020improving}. For $K>0$, let $\widetilde H(\mathcal S;\theta_0)\in[0,1]$ denote the average token-level predictive entropy under $\theta_0$, normalized by the maximum vocabulary entropy. We define
\begin{equation}
\gamma(\mathcal S)
=
\frac{1}{K}
\left[
1+\alpha\,\widetilde H(\mathcal S;\theta_0)
\right],
\qquad
\alpha\geq0,
\label{eq:evidence_weight}
\end{equation}
where $\alpha$ controls the entropy-dependent increase in anchoring strength and is selected using source-domain validation tasks before target adaptation. The unit offset preserves the PAC-Bayes-motivated $K^{-1}$ scaling when $\alpha=0$. A high entropy value means that $\theta_0$ distributes probability mass across many plausible continuations; with only a few observations, the support loss then provides less decisive evidence for how far the model should move from the transferable initialization. Accordingly, higher uncertainty strengthens the anchor without preventing adaptation altogether.

The group-wise variances $\sigma_{g(l)}^2$ allocate source-learned adaptation freedom across network depths. Within each layer, different parameter directions can have substantially different effects on the support loss. Let $\ell_i(\theta_0)$ be the token-averaged generation loss on the $i$-th support interaction. We estimate a diagonal empirical Fisher proxy at $\theta_0$ as
\begin{equation}
g_{i,l}
=
\nabla_{\theta_l}\ell_i(\theta_0),
\qquad
\widehat F_l
=
\operatorname{Diag}
\left(
\frac{1}{K}
\sum_{i=1}^{K}
g_{i,l}^{\odot 2}
\right).
\label{eq:empirical_fisher}
\end{equation}
This squared-gradient estimate provides a tractable, support-conditioned approximation to local anisotropic sensitivity. Larger entries identify directions in which the support loss is more sensitive to parameter changes, allowing the regularizer to constrain these directions more strongly while retaining greater flexibility elsewhere. We compute $\widehat F_l$ once at $\theta_0$ and keep it fixed throughout the inner-loop updates. Defining $\Delta\theta_l=\theta_l-\theta_{0,l}$, the resulting adaptation objective is
\begin{equation}
\mathcal L_{\mathrm{adapt}}(\theta;\mathcal S)
=
\widehat{\mathcal L}_{\mathcal S}(\theta)
+
\gamma(\mathcal S)
\sum_{l=1}^{L}
\frac{
\Delta\theta_l^\top
\widehat F_l
\Delta\theta_l
}{
2\sigma_{g(l)}^2
},
\label{eq:adapt_objective}
\end{equation}
where $\widehat{\mathcal L}_{\mathcal S}(\theta)$ is the average token-level generation loss on the support set. $\gamma(\mathcal S)$ controls the overall anchoring strength, $\sigma_{g(l)}^2$ allocates adaptation freedom across depth groups, and $\widehat F_l$ shapes the constraint across local parameter directions. The PAC-Bayes derivation supplies the prior-centered geometry and $K^{-1}$ dependence, while predictive entropy and empirical Fisher calibrate the deterministic surrogate to the current support evidence. Because $\widehat F_l$ is support-dependent, it is used only to reweight the deterministic adaptation surrogate and is not part of the target-support-independent prior $P_0$. Full derivations are provided in Appendix~C.

\paragraph{Episodic Meta-Optimization and Target-Domain Adaptation.}
During source-domain meta-training, each user--domain pair defines a task $\tau$ with disjoint support and query sets, $\mathcal S_\tau$ and $\mathcal Q_\tau$. Starting from $\theta_0$, the inner loop performs $M$ gradient steps on $\mathcal L_{\mathrm{adapt}}(\theta;\mathcal S_\tau)$, after which the outer loop minimizes the generation loss of the adapted parameters on $\mathcal Q_\tau$ and jointly updates $\theta_0$ and the group-wise variances $\{\sigma_g^2\}$. Because query examples are excluded from within-task adaptation, the outer objective learns a prior center and depth-wise adaptation freedoms that improve generation on new interactions after constrained adaptation, rather than merely fitting the support set. For each episode, $\widetilde H(\mathcal S_\tau;\theta_0)$ and $\{\widehat F_l\}$ are computed once at $\theta_0$ and held fixed across all inner-loop steps. Gradients are stopped only through the construction of these entropy and Fisher statistics; this does not freeze $\theta_0$, which remains an outer-loop parameter and serves as both the initialization and anchor center. The same within-task objective and update rule are used during source-domain meta-training and few-shot target-domain adaptation. Training and inference algorithms are given in Appendix~F.

\subsection{Graph-Guided Domain Prior Injection}

Cross-domain personalization requires two functionally distinct forms of conditioning: relatively stable user preference tendencies and target-domain conditions governing how those tendencies should be realized. We therefore adopt a functional, rather than semantic, factorization of personalization priors. A textual summary of the user's cross-domain history $\mathcal H_u$ specifies stable preference tendencies and is injected as a prompt to preserve semantic transparency. The domain prior instead specifies continuous realization conditions, such as information organization, discourse conventions, and response structure, and is injected as soft tokens that directly condition hidden-state computation. Their distinction concerns their roles and conditioning interfaces during generation, rather than mutually exclusive semantic content. We treat the textual user prompt as standard conditioning and focus below on constructing the domain prior from source-domain relations and injecting it through topology-preserving soft tokens.

\paragraph{Reliable Multi-View Domain Graph and Target-Conditioned Composition.}
For each source domain $d\in\mathcal D_s$, we construct three complementary observable views that describe what the domain discusses, how its interactions are conventionally organized, and what response forms are typically expected. The semantic-content view captures topics, entities, and domain-specific concepts; the interaction-pragmatics view characterizes discourse organization and expression conventions; and the response-structure view summarizes statistical output patterns such as response length and paragraph or list organization. 

Because these views contain heterogeneous feature types, each view is first processed by a geometry-compatible transformation to obtain a comparable view-specific embedding. For every domain--view pair, we estimate a reliability score that increases with effective sample size and decreases with estimation uncertainty. Each domain forms a node in a view-specific graph, and each edge combines the similarity between the corresponding embeddings with the reliability of the participating estimates. Edges whose reliability-adjusted similarity falls below a fixed threshold are removed, and normalized weighted neighborhood aggregation is applied over the remaining edges to reduce noise without propagating unreliable relations. After mapping the resulting view-specific node embeddings to probability measures on a shared feature support, we compute an entropy-regularized Wasserstein barycenter using normalized reliability weights and map the resulting measure back to a domain representation $z_d\in\mathbb R^{h_z}$, where $h_z$ is the representation dimensionality. This fusion preserves the relational geometry expressed by the heterogeneous views while reducing the influence of sparse or high-variance estimates. The resulting source-domain representations form the prior bank $\mathcal B=\{z_d:d\in\mathcal D_s\}$.

To learn how source-domain priors should be composed for an unseen domain, we adopt leave-one-domain-out training. Each source domain $d$ is treated in turn as a pseudo-target, while the remaining source domains are available as references. During each leave-one-domain-out episode, the pseudo-target node and its incident edges are masked from graph propagation and excluded from the candidate bank; its full representation is used only as the reconstruction target.

A target-evidence representation $\widehat z_d$ is constructed from the pseudo-target domain description and simulated sparse interactions. Retrieval ranks the available source nodes using their graph relations, estimation reliability, and the uncertainty of the target evidence, producing a candidate set $\mathcal N_k(d)\subseteq\mathcal D_s\setminus\{d\}$ containing at most $k$ domains and retrieval confidence scores $\omega_{dd'}$ for candidates $d'\in\mathcal N_k(d)$. Target-conditioned attention composes these candidates as
\begin{equation}
\begin{gathered}
a_{dd'}
=
\mathrm{softmax}_{d'}\!
\Bigl(
\tfrac{
  \langle W_q\hat z_d,\; W_k z_{d'}\rangle
}{\sqrt{h_a}}
+
\eta\ln(\omega_{dd'}{+}\epsilon)
\Bigr),
\\[6pt]
\bar z_d
=
\textstyle\sum_{d'\in\mathcal N_k(d)}
a_{dd'}\,z_{d'}.
\end{gathered}
\label{eq:domain_fusion}
\end{equation}
Here, $W_q$ and $W_k$ are learnable projections into an $h_a$-dimensional attention space, $\eta\geq0$ controls the contribution of retrieval confidence, and $\epsilon>0$ ensures numerical stability. The coefficient $a_{dd'}$ is the final target-conditioned contribution of candidate domain $d'$, whereas $\omega_{dd'}$ reflects its confidence before composition. Thus, retrieval determines which source domains are admissible transfer references, while attention determines how their complementary information is combined. The fusion module is trained to reconstruct the held-out representation $z_d$ from $\bar z_d$ while preserving its neighborhood relations in the source-domain graph, thereby learning a transferable composition rule rather than memorizing domain identities.

\paragraph{Topology-Preserving Soft-Token Injection.}
The fused representation $\bar z_d$ lies in an external domain space and must be converted into a condition that can participate directly in language-model computation. We use a projector $\Pi_\psi$, parameterized by $\psi$, to produce $n_T$ soft tokens:
\begin{equation}
\begin{aligned}
T_d
&=
\Pi_\psi(\bar z_d)
=
\operatorname{reshape}
\left[
W_2\varphi(W_1\bar z_d)
\right]
\in\mathbb R^{n_T\times h},\\
\mathcal L_{\mathrm{topo}}
&=
1-\operatorname{CKA}(\overline Z,\overline T)
+
\lambda_{\mathrm{attn}}
\left\|S^A-S^z\right\|_F^2.
\end{aligned}
\label{eq:soft_token_projection}
\end{equation}
Here, $W_1$ and $W_2$ are the projector parameters, $\varphi$ is its nonlinear activation, $n_T$ is the number of domain tokens, and $h$ is the hidden-state dimensionality of the language model. The matrices $\overline Z$ and $\overline T$ respectively collect the fused domain representations in a training batch and the pooled representations of their corresponding token sequences. The first term uses centered kernel alignment (CKA) to preserve the global relational geometry between these two spaces. Each entry of $S^z$ is the pairwise cosine similarity between two fused domain representations, while each entry of $S^A$ is the corresponding symmetrized cosine similarity after their soft tokens pass through the frozen query and key projections of the language model. The coefficient $\lambda_{\mathrm{attn}}\geq0$ controls this attention-space alignment, and $\|\cdot\|_F$ denotes the Frobenius norm. The second term therefore preserves source-domain relations at the actual interface through which the language model consumes the injected tokens.

The fusion module and projector are jointly optimized with the episodic query-generation loss, together with the held-out reconstruction and relation-alignment objectives. During each meta-training episode, $T_d$ is injected into both the support and query inputs of the pseudo-target task and remains fixed throughout within-task adaptation, ensuring that Meta-LoRA is trained and evaluated under the same domain condition. For an unseen target domain $d_t$, the same retrieval and composition mechanism produces $\bar z_t$, which is projected into $T_t=\Pi_\psi(\bar z_t)$ and injected together with the textual user prompt. Unlike standard retrieve-and-project pipelines, our design preserves source-domain relation structure throughout retrieval, target-conditioned composition, and projection into the language model's attention space. Feature construction and graph estimation are detailed in Appendix~D.

\paragraph{User-Level Textual Conditioning.}
A source-domain trained summarizer converts user's cross-domain history $\mathcal H_u$ into a textual preference prompt $P_u$. Trained with held-out personalization utility and cross-view consistency, it emphasizes persistent preferences while suppressing source-specific or transient patterns. We prepend $P_u$ to support and query inputs during meta-training, and compute it once before target adaptation without using current responses. The textual prompt provides interpretable user context, complementing the graph-derived domain tokens $T_d$. Source-only summarizer training is detailed in Appendix~E.

\subsection{Unified Zero- and Few-Shot Inference}
\label{sec:unified_inference}

Given a target support set $\mathcal S_t=\{(x_i,y_i)\}_{i=1}^{K}$, the target-domain description and the available support inputs $\{x_i\}_{i=1}^{K}$ are encoded in the domain representation space to construct the uncertainty-aware target-evidence representation $\widehat z_t$. Equation~\eqref{eq:domain_fusion} retrieves and composes relevant source-domain representations into $\bar z_t$, which is projected by Eq.~\eqref{eq:soft_token_projection} into $T_t$. In parallel, user's cross-domain history $\mathcal H_u$ is summarized into the textual preference prompt $P_u$. The current support responses $\{y_i\}_{i=1}^{K}$ are used only to compute the support-based adaptation loss and task statistics. The prompt and domain tokens are computed once per target episode and held fixed across all inner-loop updates.

When $K$=0, the model generates directly from $[P_u;T_t;x_t]$ using $\theta_0$. When $K$>0, the support examples are conditioned on the same prompt and domain tokens, $\widetilde H(\mathcal S_t;\theta_0)$ and $\{\widehat F_l\}$ are estimated once at $\theta_0$, and $M$ updates of $\mathcal L_{\mathrm{adapt}}$ produce the task-adapted parameters $\theta^\ast$, which are used to generate the final response from $[P_u;T_t;x_t]$. This shared conditioning interface supports zero-shot generation and few-shot evidence-calibrated refinement within the same framework.
\section{Experiments}

\begin{table*}[!h]
\centering
\small
\setlength{\tabcolsep}{4.4pt}
\renewcommand{\arraystretch}{1.15}
\resizebox{0.99\textwidth}{!}{
\begin{tabular}{l *{6}{cc} c}
\toprule
\multirow{3}{*}{Method}
& \multicolumn{2}{c}{In-domain}
& \multicolumn{10}{c}{Cross-domain}
& \multirow{3}{*}{$\Delta$WR$\downarrow$} \\
\cmidrule(lr){2-3}\cmidrule(lr){4-13}
& \multicolumn{2}{c}{Avg. over src domains}
& \multicolumn{2}{c}{Politics}
& \multicolumn{2}{c}{Music}
& \multicolumn{2}{c}{Finance}
& \multicolumn{2}{c}{Beauty}
& \multicolumn{2}{c}{Food}
& \\
\cmidrule(lr){2-3}\cmidrule(lr){4-5}\cmidrule(lr){6-7}\cmidrule(lr){8-9}\cmidrule(lr){10-11}\cmidrule(lr){12-13}
& WR$\uparrow$ & CCS$\uparrow$
& WR$\uparrow$ & CCS$\uparrow$
& WR$\uparrow$ & CCS$\uparrow$
& WR$\uparrow$ & CCS$\uparrow$
& WR$\uparrow$ & CCS$\uparrow$
& WR$\uparrow$ & CCS$\uparrow$
& \\
\midrule
SFT & \wrstd{42.6}{1.1} & 1.04 & \wrstd{32.9}{0.5} & 0.84 & \wrstd{27.1}{0.7} & 0.86 & \wrstd{23.4}{0.7} & 0.79 & \wrstd{29.4}{1.2} & 0.82 & \wrstd{26.9}{1.1} & 0.76 & 14.66 \\
LaMP & \wrstd{55.9}{1.3} & 1.19 & \wrstd{40.9}{0.6} & 0.99 & \wrstd{39.6}{0.9} & 0.97 & \wrstd{36.2}{0.5} & 0.93 & \wrstd{38.7}{0.7} & 0.98 & \wrstd{35.9}{1.0} & 0.85 & 17.64 \\
GPG & \wrstd{59.3}{0.5} & 1.29 & \wrstd{50.7}{0.7} & 1.17 & \wrstd{49.7}{1.1} & 1.18 & \wrstd{43.6}{1.0} & 1.14 & \wrstd{45.4}{0.7} & 1.16 & \wrstd{45.2}{1.0} & 1.15 & 12.38 \\
\midrule
SPT & \wrstd{60.8}{1.2} & 1.21 & \wrstd{46.4}{0.5} & 1.03 & \wrstd{45.7}{1.2} & 1.04 & \wrstd{40.6}{1.1} & 1.01 & \wrstd{47.8}{0.8} & 0.99 & \wrstd{44.5}{0.6} & 0.95 & 15.80 \\
IDL & \wrstd{62.5}{1.4} & 1.33 & \wrstd{53.0}{0.8} & 1.05 & \wrstd{51.0}{0.6} & 1.04 & \wrstd{48.0}{0.6} & 1.06 & \wrstd{50.0}{1.3} & 1.07 & \wrstd{49.0}{1.0} & 1.01 & 12.30 \\
DEP & \wrstd{68.8}{1.2} & 1.51 & \wrstd{61.8}{1.2} & 1.25 & \wrstd{58.2}{1.0} & 1.19 & \wrstd{55.0}{1.4} & 1.17 & \second{\wrstd{61.4}{0.8}} & 1.25 & \wrstd{50.4}{1.0} & 1.09 & \second{11.44} \\
\midrule
OPPU & \wrstd{69.3}{1.2} & 1.41 & \wrstd{53.4}{1.1} & 1.21 & \wrstd{51.9}{1.3} & 1.22 & \wrstd{48.1}{1.0} & 1.11 & \wrstd{50.6}{1.1} & 1.21 & \wrstd{48.3}{0.5} & 1.13 & 18.84 \\
CoPL & \wrstd{71.2}{0.7} & 1.45 & \wrstd{62.0}{0.8} & 1.24 & \wrstd{60.9}{0.6} & 1.19 & \wrstd{56.5}{0.7} & \second{1.28} & \wrstd{59.7}{0.6} & 1.18 & \wrstd{58.4}{0.8} & \second{1.16} & 11.70 \\
PROPER-GMoE & \wrstd{68.4}{1.1} & 1.31 & \wrstd{58.0}{0.8} & 1.18 & \wrstd{56.0}{0.8} & 1.17 & \wrstd{54.0}{0.7} & 1.12 & \wrstd{56.0}{0.7} & 1.13 & \wrstd{55.0}{1.3} & 1.12 & 12.60 \\
PROPER & \second{\wrstd{76.3}{1.1}} & \second{1.53} & \second{\wrstd{69.3}{1.0}} & \second{1.34} & \second{\wrstd{64.0}{0.7}} & \second{1.37} & \second{\wrstd{59.4}{1.2}} & 1.27 & \wrstd{61.0}{0.6} & \second{1.29} & \second{\wrstd{63.4}{0.8}} & 1.15 & 12.98 \\
\midrule
\rowcolor{graybg}
Ours & \best{\wrstd{79.4}{0.7}} & \best{1.64} & \best{\wrstd{76.9}{1.3}} & \best{1.48} & \best{\wrstd{71.0}{1.1}} & \best{1.41} & \best{\wrstd{73.6}{1.1}} & \best{1.39} & \best{\wrstd{75.1}{1.3}} & \best{1.32} & \best{\wrstd{69.5}{1.2}} & \best{1.41} & \best{5.98} \\
\midrule
\textit{Gains (vs. 2nd)}
& \textbf{+4.1\%}
& \textbf{+7.2\%}
& \textbf{+11.0\%}
& \textbf{+10.4\%}
& \textbf{+10.9\%}
& \textbf{+2.9\%}
& \textbf{+23.9\%}
& \textbf{+8.6\%}
& \textbf{+22.3\%}
& \textbf{+2.3\%}
& \textbf{+9.6\%}
& \textbf{+21.6\%}
& \textbf{$\downarrow$47.9\%} \\
\bottomrule
\end{tabular}}
\caption{Cross-domain 5-shot results on HiCUPID. $\Delta\mathrm{WR}\downarrow=\mathrm{WR}_{\mathrm{in}}-\operatorname{Avg}(\mathrm{WR}_{\mathrm{target}})$ measures cross-domain degradation.}
\label{tab:hicupid_main}
\vspace{-2mm}
\end{table*}

\paragraph{Experimental Setup.}
We evaluate on HiCUPID~\citep{mok-etal-2025-exploring} and LaMP-QA~\citep{salemi2025lamp}, using disjoint source domains for meta-training and held-out domains for evaluation. We use LLaMA-3-8B-Instruct as the default backbone and Qwen3-8B for cross-model evaluation. The backbone is frozen. 
Source-domain training optimizes the method-specific components. Before target adaptation, all components except the LoRA parameters are frozen. Hyperparameters are selected only on source-domain validation episodes, with identical support splits and decoding settings across methods.
Complete protocols are in Appendix~G.

\paragraph{Metrics and Baselines.}
Win Rate (WR), our primary metric, is the percentage of test instances for which the pairwise evaluator prefers the generated personalized response over the corresponding reference response. CCS aggregates utility, honesty, and personal fit, while PG (Personal Gap) measures the semantic distinction from a generic response. We define $\Delta\mathrm{WR}$ as source-domain WR minus mean target-domain WR. HiCUPID uses its official evaluator, whereas the LaMP-QA evaluator is validated against blinded human judgments. Baselines cover retrieval/text conditioning (LaMP~\citep{salemi2024lamp} and GPG~\citep{zhang2024guided}), continuous prompts (SPT~\citep{huang2024selective} and DEP~\citep{qiu2025latent}), parameter adaptation (IDL~\citep{cheng2024dialogues}, OPPU~\citep{tan2024democratizing}, PROPER~\citep{zhang2025proper}, and CoPL~\citep{choi2025copl}), inference-time personalization (BiPO~\citep{cao2024personalized}, CHAMELEON~\citep{zhang2025personalize}, OPAD~\citep{zhu2025fly}, and Proactive~\citep{zhang2025towards}), and Model-Agnostic Meta-Learning (MAML~\citep{finn2017model}). 
Where applicable, baselines use the same target support sets. All methods use matched backbones, decoding settings, and evaluation instances.
Reproducibility, limitations, and other details are in Appendix~I.

\begin{table}[h!]
\centering
\small
\setlength{\tabcolsep}{4.2pt}
\renewcommand{\arraystretch}{1.15}
\resizebox{\columnwidth}{!}{%
\begin{tabular}{l ccc ccc c}
\toprule
\multirow{2}{*}{\textbf{Method}}
& \multicolumn{6}{c}{\textbf{Cross-domain}}
& \multirow{2}{*}{$\boldsymbol{\Delta}$\textbf{WR}} \\
\cmidrule(lr){2-7}
& \multicolumn{3}{c}{\textbf{Target: parenting}}
& \multicolumn{3}{c}{\textbf{Target: interpersonal}}
& \\
\cmidrule(lr){2-4}\cmidrule(lr){5-7}
& WR & CCS & PG
& WR & CCS & PG
& \\
\midrule
SFT
& \wrstd{46.2}{0.9} & 1.094 & 0.033
& \wrstd{44.7}{1.1} & 1.112 & 0.044
& 15.25 \\
LaMP
& \wrstd{50.8}{1.0} & 1.182 & 0.036
& \wrstd{49.1}{0.8} & 1.201 & 0.036
& 15.45 \\
\midrule
OPPU
& \wrstd{54.1}{1.2} & 1.214 & 0.045
& \wrstd{53.4}{1.3} & 1.226 & 0.043
& 14.85 \\
BiPO
& \wrstd{55.3}{1.1} & 1.302 & 0.049
& \wrstd{54.8}{1.0} & 1.334 & \second{0.052}
& 15.85 \\
CHAMELEON
& \wrstd{56.8}{0.7} & 1.418 & \second{0.057}
& \wrstd{55.2}{1.1} & \second{1.447} & 0.049
& 15.60 \\
PROPER
& \second{\wrstd{59.6}{0.8}} & \second{1.456} & 0.056
& \second{\wrstd{59.1}{1.2}} & 1.401 & 0.051
& \second{12.75} \\
\midrule
\rowcolor{graybg}
\textbf{Ours}
& \textbf{\wrstd{69.9}{0.7}} & \textbf{1.528} & \textbf{0.066}
& \textbf{\wrstd{67.8}{0.9}} & \textbf{1.571} & \textbf{0.061}
& \textbf{5.45} \\
\midrule
\textit{Gains (vs. 2nd)}
& \textbf{+17.3\%} & \textbf{+4.9\%} & \textbf{+15.8\%}
& \textbf{+14.7\%} & \textbf{+8.6\%} & \textbf{+17.3\%}
& \textbf{$\downarrow$57.3\%} \\
\bottomrule
\end{tabular}%
}
\caption{Performance on LaMP-QA under the cross-domain 5-shot personalization setting. WR is reported as mean$\pm$std.}
\label{tab:lamp_main}
\end{table}

\begin{table}[!ht]
    \centering
    \small
    \setlength{\tabcolsep}{4.0pt}
    \renewcommand{\arraystretch}{1.12}
    \resizebox{\columnwidth}{!}{%
    \begin{tabular}{l ccccc c}
        \toprule
        \multirow{2}{*}{\textbf{Method}}
        & \multicolumn{5}{c}{\textbf{Cross-domain Avg.}}
        & \multirow{2}{*}{$\boldsymbol{\Delta}$\textbf{WR}} \\
        \cmidrule(lr){2-6}
        & WR & PG & Util & Hon & Fit & \\
        \midrule
        LoRA & 24.8{\scriptsize$\pm$1.4} & 0.004 & 0.92 & 1.02 & 0.70 & 24.5 \\
        DEP            & 50.7{\scriptsize$\pm$1.2} & 0.033 & 1.04 & 1.15 & 0.98 & 17.6 \\
        OPPU           & 47.1{\scriptsize$\pm$1.3} & 0.039 & 1.05 & 1.16 & 0.93 & 22.3 \\
        PROPER         & 58.5{\scriptsize$\pm$1.1} & 0.051 & 1.32 & 1.27 & 1.09 & 19.2 \\
        \midrule
        \rowcolor{graybg}
        \textbf{Ours}  & \textbf{70.9}{\scriptsize$\pm$0.8} & \textbf{0.073} & \textbf{1.56} & \textbf{1.52} & \textbf{1.33} & \textbf{6.7} \\
        \midrule
        \textit{Gains (vs. 2nd)} & \textbf{+21.2\%} & \textbf{+43.1\%} & \textbf{+18.2\%} & \textbf{+20.8\%} & \textbf{+22.0\%} & \textbf{$\downarrow$61.9\%} \\
        \bottomrule
    \end{tabular}%
    }
    \caption{Performance on the \texttt{Qwen3-8B} backbone.}
    \label{tab:qwen_results}
    \vspace{-2mm}
\end{table}

\begin{table}[t]
    \centering
    \resizebox{0.95\columnwidth}{!}{%
    \begin{tabular}{lcccc}
        \toprule
        \textbf{Method} & \textbf{Win Rate} & \textbf{PG} & \textbf{Personal Fit} \\
        \midrule
        Vanilla       & 5.1  & 0.004  & 0.37 \\
        SFT        & 11.1 & 0.013  & 0.47 \\
        Proactive      & 15.4 & 0.019  & 0.58 \\
        PROPER         & 19.9 & 0.024  & 0.79 \\
        OPAD           & 20.7 & 0.021  & 0.61 \\
        \midrule
        \rowcolor{graybg} \textbf{Ours} & \textbf{43.5} & \textbf{0.027} &  \textbf{1.12} \\
        \midrule
        \textit{Gains (vs. 2nd)} & \textbf{+110.2\%} & \textbf{+12.5\%} &  \textbf{+41.8\%} \\
        \bottomrule
    \end{tabular}%
    }
    \caption{Performance in the \textbf{Unseen User} scenario.}
    \label{tab:unseen_user}
    \vspace{-3mm}
\end{table}
\paragraph{Main Results.}
Table~\ref{tab:hicupid_main} shows that our method achieves the strongest 5-shot cross-domain personalization performance on HiCUPID. It obtains the best WR and CCS on all five held-out target domains. More importantly, it reduces $\Delta\mathrm{WR}$ to $5.98$, corresponding to a $47.9\%$ relative reduction in cross-domain degradation. These results indicate that the improvement is not merely due to stronger source-domain fitting, but reflects substantially more stable transfer of personalized behavior to unseen domains.

The advantage also generalizes across datasets and backbone architectures. On LaMP-QA, Table~\ref{tab:lamp_main} shows that our method improves WR over PROPER by $17.3\%$ on Parenting and $14.7\%$ on Interpersonal, while reducing $\Delta\mathrm{WR}$ from $12.75$ to $5.45$. On Qwen3-8B, Table~\ref{tab:qwen_results} reports a cross-domain WR of $70.9$, a $21.2\%$ relative improvement over PROPER, together with the smallest degradation of $6.7$. The consistent gains demonstrate that the framework is not tied to a particular task format or model architecture.

The unseen-user setting is more challenging because the target user has no personal interaction history, leaving the model without direct user evidence. As shown in Table~\ref{tab:unseen_user}, our method achieves a WR of $43.5$, more than twice that of the strongest baseline. In this extreme cold-start regime, the graph-guided domain-prior construction supplies structured target-domain conditions from related source domains, helping the model avoid generic responses.

\begin{table}[t]
\centering
\small
\setlength{\tabcolsep}{2.6pt}
\renewcommand{\arraystretch}{1.08}
\resizebox{\columnwidth}{!}{
\begin{tabular}{lcccccc}
\toprule
\multirow{2}{*}{Variant}
& \multicolumn{3}{c}{0-shot}
& \multicolumn{3}{c}{5-shot} \\
\cmidrule(lr){2-4}\cmidrule(lr){5-7}
& WR$\uparrow$ & CCS$\uparrow$ & PG$\uparrow$
& WR$\uparrow$ & CCS$\uparrow$ & PG$\uparrow$ \\
\midrule
Full Model
& \best{\wrstd{59.5}{1.1}} & \best{1.29} & \best{.053}
& \best{\wrstd{73.2}{1.3}} & \best{1.40} & \best{.091} \\
\midrule
\multicolumn{7}{l}{\textit{Core-design attribution}} \\
Textual conditioning only
& \wrstd{55.8}{2.4} & 1.21 & .043
& \wrstd{59.3}{1.9} & 1.26 & .057 \\
+ Domain soft tokens
& \wrstd{59.0}{1.4} & 1.27 & .050
& \wrstd{63.4}{1.7} & 1.31 & .066 \\
+ Vanilla Meta-LoRA
& \wrstd{59.2}{1.2} & 1.28 & .051
& \wrstd{68.4}{1.6} & 1.34 & .076 \\
\midrule
\multicolumn{7}{l}{\textit{Evidence-calibrated adaptation}} \\
w/o $K^{-1}$ scaling
& \wrstd{55.9}{2.1} & 1.28 & .051
& \wrstd{69.6}{1.7} & 1.33 & .078 \\
w/o entropy calibration
& \wrstd{57.6}{1.2} & 1.28 & .052
& \wrstd{70.8}{1.5} & 1.35 & .082 \\
Single shared variance
& \wrstd{59.3}{1.1} & 1.29 & .052
& \wrstd{71.7}{0.6} & 1.37 & .085 \\
w/o Fisher weighting
& \wrstd{58.2}{0.9} & 1.29 & .053
& \wrstd{72.2}{2.3} & 1.38 & .087 \\
\midrule
\multicolumn{7}{l}{\textit{Graph-guided domain injection}} \\
w/o graph reliability
& \wrstd{57.8}{1.4} & 1.24 & .047
& \wrstd{70.3}{1.7} & 1.34 & .080 \\
w/o topology preservation
& \wrstd{58.4}{1.3} & 1.25 & .049
& \wrstd{71.0}{0.8} & 1.36 & .084 \\
\bottomrule
\end{tabular}}
\caption{Ablation results averaged over five target domains.}
\label{tab:ablation}
\vspace{-4mm}
\end{table}

\paragraph{Ablation Study.}
Table~\ref{tab:ablation} provides cumulative attribution and component-wise ablations. Starting from textual conditioning, adding domain soft tokens improves 5-shot WR from $59.3$ to $63.4$, vanilla Meta-LoRA further raises it to $68.4$, and evidence-calibrated adaptation reaches $73.2$. The corresponding 0-shot gain from $55.8$ to $59.5$ shows that domain conditioning also improves transfer before target updates. Among its components, removing $K^{-1}$ scaling causes the largest degradation. Removing graph reliability or topology preservation also degrades both regimes, confirming that the improvement does not arise from inserting arbitrary soft tokens. Because each ablated variant is meta-trained from scratch, changes to the inner-loop objective can also alter the learned initialization $\theta_0$ and thus affect zero-shot performance. Exact ablation definitions and additional attribution are in Appendix~H.

\begin{figure}[t]
    \centering
    \includegraphics[width=0.85\linewidth]{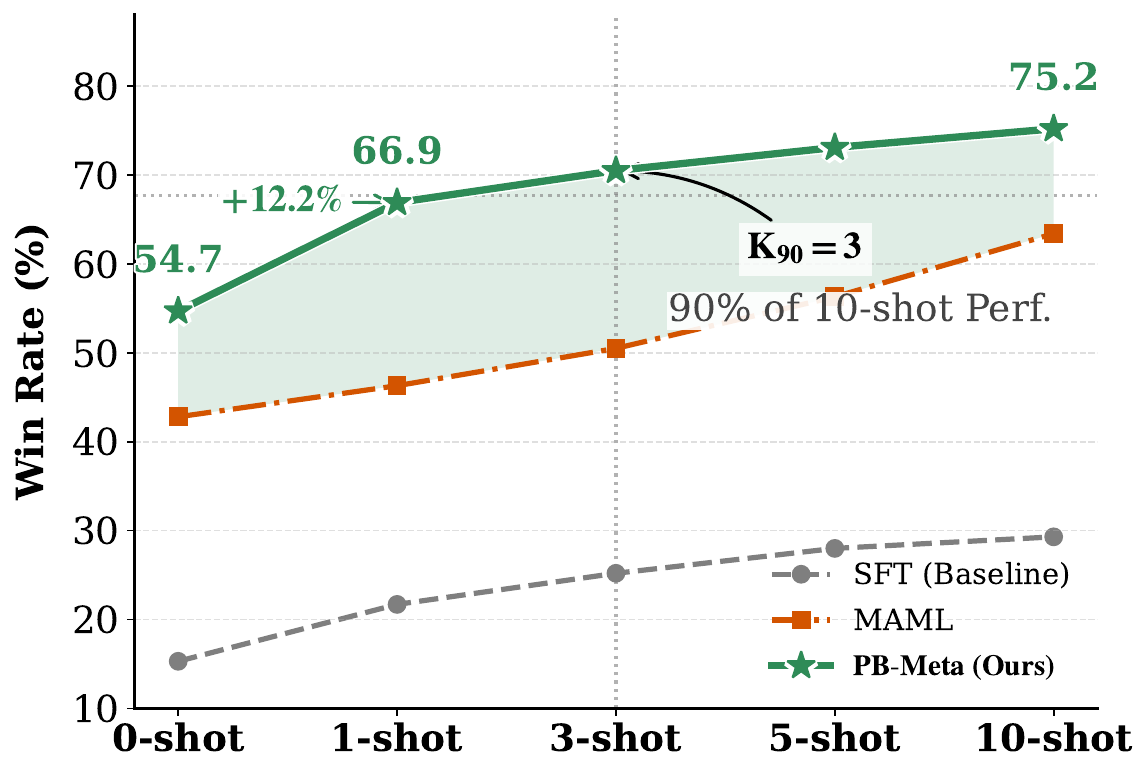}
    \caption{Sample efficiency across support sizes. }
    \label{fig:sample_efficiency}
\end{figure}

\paragraph{Sample Efficiency and Adaptation Analysis.}
Figure~\ref{fig:sample_efficiency} compares SFT, MAML, and our method across different support sizes. Our method achieves the highest WR throughout $K\in\{0,1,3,5,10\}$. Its WR increases from $54.7$ in the zero-shot setting to $66.9$ with a single target-domain interaction, and reaches approximately $90\%$ of its 10-shot performance with only three interactions. These results indicate that episodic meta-learning provides an initialization capable of rapidly exploiting sparse target-user evidence, while controlled adaptation mitigates unstable updates.

\paragraph{Case Study.}
Table~\ref{tab:case_study} presents an illustrative example. Given the same query and source-domain history, MAML captures a coarse domain-relevant interest, whereas our method integrates multiple history-supported preferences into a coherent response.

\begin{table}[t]
\centering
\small
\setlength{\tabcolsep}{2.5pt}
\renewcommand{\arraystretch}{1.05}
\begin{tabular}{@{}p{0.19\columnwidth}p{0.80\columnwidth}@{}}
\toprule
\textbf{Item} & \textbf{Content} (Target domain: Food) \\
\midrule
Query & \textit{How to relax after a busy day?} \\
Preference cues & Bubble-tea preference; interest in history and urban planning; detail orientation. \\
MAML & ``Reading about \textbf{urban design} could be a relaxing hobby.'' \\
Ours & ``Considering your past preferences, we suggest you walk through a \textbf{historic neighborhood}, pair it with a \textbf{history} read, and end with your favorite \textbf{bubble tea}.'' \\
\bottomrule
\end{tabular}
\caption{Illustrative cross-domain case. The cues are not provided as oracle inputs and are shown only for analysis.}
\label{tab:case_study}
\vspace{-4mm}
\end{table}

\paragraph{Extreme Domain-Shift Robustness.}
Using an independent external encoder, we represent each domain by the $\ell_2$-normalized mean embedding $\mu_d$ of an equal-sized held-out set of interactions. We define target--source proximity as $s_t=\max_{d_s\in\mathcal D_s}\cos(\mu_{d_t},\mu_{d_s})$ and domain shift as $1-s_t$. LaMP-QA Philosophy and HiCUPID Food have the lowest target--source proximities ($0.156/0.302$). Our method improves WR/CCS over the strongest baseline by $15.4\%/13.9\%$ and $9.6\%/21.6\%$, respectively, showing that its gains persist under the largest observed shifts. Experimental details and supplementary analyses (latency, metric reliability, topology preservation, etc.) are provided in Appendix~H.
\section{Conclusion}

We study cross-domain zero- and few-shot LLM personalization under sparse target-domain evidence. Our framework combines evidence-calibrated Meta-LoRA, which regulates adaptation through PAC-Bayes-guided anchoring, with graph-guided domain injection that preserves domain relations through soft tokens while retaining interpretable textual user prompts. Experiments show consistent gains across domains, backbones, support sizes, and severe cold-start settings. Ablation and structural analyses further validate the contributions of adaptive regularization and topology-preserving domain transfer.

\bibliography{aaai2027_arxiv}

\clearpage
\setcounter{secnumdepth}{2}
\appendix
\section*{Appendix}

\section{Supplementary Contents}
\label{app:contents}

\begin{description}
\item[Appendix B.] Task definition, source episodes, information boundaries, and notation.
\item[Appendix C.] PAC-Bayes assumptions, derivations, deterministic surrogate, and computational cost.
\item[Appendix D.] Multi-view domain features, reliability graph, Wasserstein fusion, retrieval, and topology-preserving projection.
\item[Appendix E.] Source-only user-preference summarizer training and target-time use.
\item[Appendix F.] Complete source-training and zero-/few-shot inference algorithms.
\item[Appendix G.] Datasets, baselines, metrics, evaluators, hyperparameters, and statistical protocol.
\item[Appendix H.] Ablations, cross-backbone results, sample efficiency, optimization stability, topology, attention attribution, domain shift, latency, evaluator agreement, sensitivity, and case study.
\item[Appendix I.] Reproducibility package, limitations, privacy, and safety.
\end{description}

\section{Task Definition and Information Boundaries}
\label{app:task}

\subsection{Cross-Domain Personalization Task}
\label{app:task_definition}

Let $\mathcal D_s$ be the set of source domains available during training and let $d_t\notin\mathcal D_s$ be a held-out target domain. For a target user $u$, $\mathcal H_u$ contains only pre-target interactions from source domains and may be empty in the unseen-user evaluation. The optional target support set is $\mathcal S_t=\{(x_i,y_i)\}_{i=1}^{K}$, where $K=0$ denotes zero-shot inference and $K>0$ denotes few-shot adaptation. The test query $x_t$ and its response $y_t$ are disjoint from $\mathcal S_t$. The target-domain description is public task metadata rather than user supervision.

The framework constructs a textual user prompt $P_u$, a target-domain soft-token sequence $T_t$, and, when $K>0$, adapted low-rank parameters~\citep{hu2022lora} $\theta_t^\ast$. The generated response is
\begin{equation}
\widehat y_t=
\begin{cases}
G([P_u;T_t;x_t];\theta_0), & K=0,\\
G([P_u;T_t;x_t];\theta_t^\ast), & K>0,
\end{cases}
\label{eq:app_prediction}
\end{equation}
where the backbone language-model parameters are frozen and only LoRA parameters are changed during target adaptation.

\subsection{Functional Factorization and Claim Boundary}
\label{app:functional_factorization}

The user/domain decomposition is functional rather than an attribute-wise hard partition. The textual prompt summarizes tendencies that are stable across the observed history, whereas the domain tokens specify how those tendencies should be realized under the current domain's terminology, discourse structure, and pragmatic conventions. Consequently, an attribute such as tone or language is not assigned permanently to one channel. For example, a preference for empathetic explanations can be retained in $P_u$, while its wording, degree of subjectivity, and evidence style are modulated by $T_t$ in a technical or social domain. Both channels condition the same generator and are optimized by held-out source-task performance. We therefore claim complementary conditioning roles, not identifiable or statistically independent latent factors.

\subsection{Source-Domain Episodic Tasks}
\label{app:source_tasks}

Each source task $\tau$ is a user--domain pair. Following episodic gradient-based meta-learning~\citep{pmlr-v70-finn17a}, its interactions are partitioned into disjoint support and query sets, $\mathcal S_\tau$ and $\mathcal Q_\tau$. The support set simulates target-time adaptation, whereas the query set estimates performance on a new interaction from the same user--domain task. No query response is used by the inner loop. Support sizes are sampled from the same range evaluated at target time so that the meta-objective learns an initialization and regularization geometry across different evidence regimes.

\subsection{Information-Access Matrix}
\label{app:information_boundary}

Table~\ref{tab:information_boundary} makes the access boundary explicit. In particular, target support responses supervise only the few-shot LoRA update and its support-conditioned entropy and Fisher statistics. They are not used to construct either personalization prior. This separation prevents the prior bank, user prompt, and target-domain tokens from encoding the answers that are later used as adaptation supervision.

\begin{table*}[t]
\centering
\small
\setlength{\tabcolsep}{5.0pt}
\renewcommand{\arraystretch}{1.12}
\resizebox{\textwidth}{!}{%
\begin{tabular}{lccccc}
\toprule
Component & Source interactions & Pre-target user history $\mathcal H_u$ & Target description & Support inputs $\{x_i\}$ & Support responses $\{y_i\}$ \\
\midrule
Meta-initialization and group variances & Yes & No & No & No & No \\
Source-domain prior bank & Yes & No & No & No & No \\
Preference summarizer training & Yes & No & No & No & No \\
Target user prompt $P_u$ & Frozen summarizer & Yes & No & No & No \\
Target evidence $\widehat z_t$ and tokens $T_t$ & Frozen domain modules & No & Yes & Yes, if $K>0$ & No \\
Entropy and Fisher statistics & Frozen $\theta_0$ & Through conditioning & Through $T_t$ & Yes & Yes \\
Target LoRA adaptation & Frozen $\theta_0$ & Through $P_u$ & Through $T_t$ & Yes & Yes \\
Test generation & Frozen modules & Through $P_u$ & Through $T_t$ & Indirectly via adaptation & No test response \\
\bottomrule
\end{tabular}%
}
\caption{Information available to each component. ``Source only'' means that the component is learned before the target support set is observed.}
\label{tab:information_boundary}
\end{table*}

\subsection{Offline and Online States}
\label{app:frozen_states}

The backbone, preference summarizer, source-domain bank, graph encoder, retrieval and composition modules, soft-token projector, meta-initialization, and learned group variances are fixed before a target episode begins. At target time, $P_u$ and $T_t$ are computed once and held fixed across all inner-loop steps. The entropy and diagonal Fisher proxy are also computed once at $\theta_0$ and treated as stop-gradient task statistics. Only the target copy of the LoRA parameters is updated, and that copy is discarded after the episode. Consequently, one target user's adaptation cannot modify the shared model or another user's state.

\subsection{Notation}
\label{app:notation}

\begin{table}[t]
\centering
\small
\setlength{\tabcolsep}{4.0pt}
\renewcommand{\arraystretch}{1.08}
\resizebox{\columnwidth}{!}{%
\begin{tabular}{ll}
\toprule
Symbol & Meaning \\
\midrule
$\mathcal D_s,d_t$ & source-domain set and held-out target domain \\
$\mathcal H_u$ & target user's pre-target source-domain history \\
$\mathcal S_t,K$ & target support set and number of interactions \\
$P_u,T_t$ & textual user prompt and target-domain soft tokens \\
$\theta_0,\theta$ & meta-initialized and current LoRA parameters \\
$g(l)$ & depth group containing LoRA layer $l$ \\
$\sigma_g^2$ & source-learned variance for depth group $g$ \\
$\widehat F_l$ & support-conditioned diagonal Fisher proxy at $\theta_0$ \\
$z_d,\bar z_d$ & source-domain representation and composed representation \\
$\widehat z_t$ & target-evidence representation \\
$\mathcal B$ & source-domain representation bank \\
\bottomrule
\end{tabular}%
}
\caption{Symbols used repeatedly in the supplement.}
\label{tab:notation}
\end{table}

\section{PAC-Bayes-Regularized Meta-LoRA}
\label{app:pac}

PAC-Bayes meta-learning permits source tasks to determine a task-transfer prior before a new task is observed~\citep{riou2023bayes}. Our construction specializes this principle to LoRA personalization: the episodically learned initialization is the prior mean, depth-group variances encode source-learned adaptation freedom, and target support size controls the deterministic anchor derived below.

\subsection{Bounded Analysis Loss}
\label{app:bounded_loss}

The training loss is the token-averaged negative log-likelihood (NLL), which is nonnegative but unbounded. The PAC-Bayes argument instead requires a bounded loss~\citep{mcallester1999some,maurer2004note,seeger2002pac}. Let $\ell_{\mathrm{nll}}(z;\theta)$ be the token-averaged NLL for interaction $z$, and define the analysis-only loss
\begin{equation}
\ell_b(z;\theta)=1-\exp\!\left(-\ell_{\mathrm{nll}}(z;\theta)/c\right)\in[0,1],
\label{eq:app_bounded_loss}
\end{equation}
where $c>0$ is fixed using source-domain validation data before any target episode. This mapping is monotone, so it preserves the ordering of example losses, but it is not used by the adaptation optimizer. All reported model updates continue to minimize the original NLL.

\subsection{PAC-Bayes Statement and Assumptions}
\label{app:pac_bound}

\paragraph{Assumptions.}
Conditioned on the complete source-training procedure, the prior $P_0$ is independent of the target support set. The $K\geq1$ target support interactions are sampled independently from the target user--domain distribution, and the bounded loss in Eq.~\eqref{eq:app_bounded_loss} is fixed before observing them. The posterior may depend on the support set.

\paragraph{Proposition 1.}
For any target-support-independent prior $P_0$, with probability at least $1-\delta$ over a support set of size $K$, every posterior $Q$ satisfies
\begin{equation}
R(Q)\leq \widehat R_{\mathcal S}(Q)+
\sqrt{\frac{\operatorname{KL}(Q\Vert P_0)+\ln((K+1)/\delta)}{2K}}.
\label{eq:app_pac_bound}
\end{equation}

\paragraph{Derivation.}
A standard PAC-Bayes change-of-measure argument for bounded losses yields the square-root form used in the main paper~\citep{maurer2004note}
\begin{equation}
R(Q)-\widehat R_{\mathcal S}(Q)
\leq
\sqrt{\frac{\operatorname{KL}(Q\Vert P_0)+\ln((K+1)/\delta)}{2K}},
\label{eq:app_kl_bound}
\end{equation}
simultaneously for all $Q$. Equation~\eqref{eq:app_pac_bound} is therefore a restatement of Eq.~\eqref{eq:app_kl_bound}; no stronger guarantee is introduced in the supplement.

\subsection{Closed-Form Gaussian KL}
\label{app:gaussian_kl}

For layer $l$, the prior and posterior have identical isotropic covariance:
\begin{equation}
P_l=\mathcal N(\theta_{0,l},\sigma_{g(l)}^2I),
\qquad
Q_l=\mathcal N(\theta_l,\sigma_{g(l)}^2I).
\end{equation}
The KL between two $d_l$-dimensional Gaussians is
\begin{align}
\operatorname{KL}(Q_l\Vert P_l)
&=\frac12\left[
\operatorname{tr}(\Sigma_l^{-1}\Sigma_l)-d_l
+\ln\frac{\det\Sigma_l}{\det\Sigma_l}
\right.\nonumber\\[-2pt]
&\hspace{24mm}\left.
+(\theta_{0,l}-\theta_l)^\top
\Sigma_l^{-1}(\theta_{0,l}-\theta_l)
\right]\nonumber\\
&=\frac{\|\theta_l-\theta_{0,l}\|_2^2}{2\sigma_{g(l)}^2}.
\label{eq:app_layer_kl}
\end{align}
Independence across layers makes the joint KL additive:
\begin{equation}
\operatorname{KL}(Q_\theta\Vert P_0)
=\sum_{l=1}^{L}\frac{\|\theta_l-\theta_{0,l}\|_2^2}{2\sigma_{g(l)}^2}.
\label{eq:app_joint_kl}
\end{equation}
No covariance, trace, determinant, or dimensionality term remains because the prior and posterior share the same covariance.

\subsection{From the Square Root to a \(K^{-1}\)-Scaled Surrogate}
\label{app:linear_surrogate}

Let $A_\theta=\operatorname{KL}(Q_\theta\Vert P_0)+\ln((K+1)/\delta)$. For any $\beta>0$ and $x\geq0$, $(x-\beta)^2\geq0$ implies $x\leq x^2/(2\beta)+\beta/2$. Substituting $x=\sqrt{A_\theta/(2K)}$ gives
\begin{equation}
\sqrt{\frac{A_\theta}{2K}}
\leq
\frac{A_\theta}{4\beta K}+\frac{\beta}{2}.
\label{eq:app_linearization}
\end{equation}
For fixed $K$, $\delta$, and $\beta$, the logarithmic term and $\beta/2$ do not depend on $\theta$. The remaining parameter-dependent contribution is
\begin{equation}
\frac{1}{4\beta K}\operatorname{KL}(Q_\theta\Vert P_0),
\label{eq:app_k_scaling}
\end{equation}
which motivates an anchor whose weight is proportional to $K^{-1}$. The constant $1/(4\beta)$ is absorbed into the source-validated regularization coefficient.

\paragraph{Scope of the guarantee.}
Equation~\eqref{eq:app_pac_bound} applies to the randomized predictor obtained by sampling from $Q_\theta$. Deployment uses the posterior mean $\theta$. We therefore use the bound to select the prior center, quadratic complexity geometry, and support-size dependence of a deterministic surrogate. We do not claim that Eq.~\eqref{eq:app_pac_bound} is itself a bound on deterministic mean-parameter generation.

\paragraph{Status and empirical coverage of the theoretical claims.}
Proposition~1 is the only formal generalization statement made by this work; it is a specialization of a standard PAC-Bayes result rather than a new PAC-Bayes theorem. Its assumptions, change-of-measure step, Pinsker reduction, Gaussian KL calculation, and linearization are given above. The subsequent entropy calibration, Fisher reweighting, and depth-group variances define a deterministic optimization surrogate and are not presented as additional generalization guarantees. Their observable implications are tested separately: the $K^{-1}$ and entropy ablations test evidence-dependent anchoring, the Fisher and shared-variance ablations test the proposed geometry, and the displacement analysis tests whether sparse-support updates remain closer to $\theta_0$. Thus, the theoretical statement is fully derived, while the method-specific consequences are evaluated empirically without conflating empirical support with proof of a deterministic risk bound.

\subsection{Predictive-Entropy Calibration}
\label{app:entropy}

Let $\mathcal I(\mathcal S)$ be the set of valid target-token positions in the support responses under teacher forcing, and let $p_{\theta_0}(v\mid c_j)$ be the next-token distribution at position $j$ given its conditioned prefix $c_j=[P_u;T_t;x_i;y_{i,<j}]$. With vocabulary $\mathcal V$, the normalized predictive entropy is
\begin{equation}
\begin{aligned}
\widetilde H(\mathcal S;\theta_0)
&=
\frac{1}{|\mathcal I(\mathcal S)|}
\sum_{j\in\mathcal I(\mathcal S)}
\frac{H(p_{\theta_0}(\cdot\mid c_j))}{\ln|\mathcal V|},
\\
H(p_{\theta_0}(\cdot\mid c_j))
&=
-\sum_{v\in\mathcal V}
p_{\theta_0}(v\mid c_j)\ln p_{\theta_0}(v\mid c_j).
\end{aligned}
\label{eq:app_entropy}
\end{equation}
Thus $\widetilde H\in[0,1]$. Entropy is a model-relative uncertainty statistic, not an intrinsic measure of data quality, consistent with uncertainty-aware calibration~\citep{krishnan2020improving}. The evidence weight is
\begin{equation}
\gamma(\mathcal S)=\frac{1+\alpha\widetilde H(\mathcal S;\theta_0)}{K},
\label{eq:app_gamma}
\end{equation}
where $\alpha\geq0$ is selected on source-domain validation episodes. The unit term retains the PAC-Bayes-motivated $K^{-1}$ dependence, and the entropy term increases anchoring when the meta-initialized predictor is diffuse on the observed responses.

\subsection{Diagonal Empirical Fisher Proxy}
\label{app:fisher}

For support interaction $i$, let $\ell_i(\theta_0)$ be its token-averaged NLL and define $g_{i,l}=\nabla_{\theta_l}\ell_i(\theta_0)$. The diagonal squared-gradient estimate is
\begin{equation}
\widehat F_l=
\diag\!\left(\frac1K\sum_{i=1}^{K}g_{i,l}^{\odot2}\right).
\label{eq:app_fisher}
\end{equation}
This is a tractable empirical Fisher proxy rather than the exact population Fisher. Squared-gradient curvature proxies are widely used to preserve parameters that are locally important to predictive behavior~\citep{kirkpatrick2017overcoming}. Its role here is operational: coordinates with larger support-loss sensitivity receive larger penalties. It is estimated once at $\theta_0$, detached from the meta-gradient graph, and held fixed during the $M$ inner-loop steps. Because $\widehat F_l$ depends on $\mathcal S$, it is not part of $P_0$ and cannot be used to justify prior independence. It only reweights the deterministic surrogate after the PAC-Bayes geometry has been derived.

\subsection{Source-Learned Depth-Group Variances}
\label{app:variances}

To ensure positivity and avoid degenerate variances, each unconstrained source-trained scalar $\rho_g$ is mapped to
\begin{equation}
\sigma_g^2=\sigma_{\min}^2+
\left(\sigma_{\max}^2-\sigma_{\min}^2\right)\operatorname{sigmoid}(\rho_g),
\label{eq:app_variance_parameterization}
\end{equation}
where the bounds and depth partition are selected using source validation only. All LoRA layers within a depth group share one variance. The variances receive outer-loop gradients through the adapted query loss, but they are not updated inside a task and are frozen before target adaptation. A smaller learned variance makes displacement in that depth group more costly; a larger variance permits greater task-specific movement.

\subsection{Deterministic Adaptation and Meta-Objective}
\label{app:meta_objective}

Combining Eqs.~\eqref{eq:app_gamma} and \eqref{eq:app_fisher}, the target-time objective is
\begin{equation}
\mathcal L_{\mathrm{adapt}}(\theta;\mathcal S)
=
\widehat{\mathcal L}_{\mathcal S}(\theta)
+
\gamma(\mathcal S)
\sum_{l=1}^{L}
\frac{(\theta_l-\theta_{0,l})^\top\widehat F_l(\theta_l-\theta_{0,l})}
{2\sigma_{g(l)}^2}.
\label{eq:app_adapt}
\end{equation}
For source task $\tau$, initialize $\theta_\tau^{(0)}=\theta_0$ and update
\begin{equation}
\begin{aligned}
\theta_\tau^{(m+1)}
&=
\theta_\tau^{(m)}
-\eta_{\mathrm{in}}
\nabla_\theta\mathcal L_{\mathrm{adapt}}
\!\left(\theta_\tau^{(m)};\mathcal S_\tau\right),
\\[-2pt]
&\hspace{28mm}m=0,\ldots,M-1.
\end{aligned}
\label{eq:app_inner_update}
\end{equation}
The outer objective is
\begin{equation}
\begin{aligned}
\mathcal L_{\mathrm{meta}}
&=
\mathbb E_{\tau}
\left[
\widehat{\mathcal L}_{\mathcal Q_\tau}
\!\left(\theta_\tau^{(M)};P_\tau,T_{d(\tau)}\right)
\right]
\\
&\quad
+\lambda_{\mathrm{comp}}\mathcal L_{\mathrm{comp}}
+\lambda_{\mathrm{topo}}\mathcal L_{\mathrm{topo}}.
\end{aligned}
\label{eq:app_outer_objective}
\end{equation}
The latter two terms train source-only domain composition and the topology-preserving projector. The query-generation term propagates through the inner updates to $\theta_0$ and the variance parameters. The preference summarizer is optimized separately with Eq.~\eqref{eq:app_grpo_loss} before episodic training and is then fixed. The implementation uses the same declared first- or second-order differentiation setting for every meta-learning baseline.

\subsection{Computational and Storage Cost}
\label{app:pac_cost}

The entropy calculation reuses the support forward pass. The Fisher proxy requires per-example LoRA gradients once at $\theta_0$ and stores one diagonal value per adapted parameter. Thereafter, each regularization evaluation is an element-wise multiply and reduction. For $p_{\mathrm{LoRA}}$ adapted parameters, the additional storage is $O(p_{\mathrm{LoRA}})$ and the additional arithmetic per inner step is $O(p_{\mathrm{LoRA}})$. No full Fisher matrix, Hessian, or target-specific backbone copy is stored.

\section{Reliable Domain Prior Construction}
\label{app:domain}

\subsection{Three Observable Views}
\label{app:domain_views}

The domain representation is deliberately based on observable aggregate evidence rather than manually assigned semantic attributes.

\begin{itemize}
\item \textbf{Semantic-content view.} Topic and entity distributions, intent frequencies, and fixed-encoder keyword representations summarize what is discussed.
\item \textbf{Interaction-pragmatics view.} Question type, instruction versus discussion ratios, terminology density, discourse markers, and organization cues summarize how interactions are conventionally expressed.
\item \textbf{Response-structure view.} Token and sentence length, paragraph count, list frequency, formatting patterns, and related aggregate statistics summarize the form expected of responses.
\end{itemize}

The views are not assumed to be semantically disjoint. They are operational groupings that expose complementary statistics and permit view-specific geometry and reliability treatment.

\subsection{Geometry-Compatible Preprocessing}
\label{app:geometry}

For a simplex-valued feature $p_i^{(v)}$, reliability shrinkage is first applied:
\begin{equation}
p_{i,\mathrm{sh}}^{(v)}
=r_i^{(v)}p_i^{(v)}
+(1-r_i^{(v)})\bar p^{(v)}.
\label{eq:app_shrinkage}
\end{equation}
After $\epsilon$ smoothing, the centered log-ratio transform maps it to Aitchison Euclidean coordinates~\citep{aitchison1986statistical}:
\begin{equation}
\operatorname{clr}(p)
=
\ln p-\frac{1}{\dim(p)}\bm 1\bm 1^\top\ln p.
\label{eq:app_clr}
\end{equation}
Proportion-valued features use a clipped logit followed by median/MAD standardization. Positive scalar features use $\ln(1+x)$ followed by the same robust standardization. Semantic vectors are $\ell_2$ normalized before similarity calculation. Each processed view is mapped by a view-specific encoder into a shared dimensionality, but view statistics are not concatenated before their geometries are normalized.

\subsection{Reliability Estimation and Graph Construction}
\label{app:reliable_graph}

For domain $d$ and view $v$, let $n_d^{(v)}$ be the effective sample size and let $u_d^{(v)}$ summarize estimation variance after bounded robust preprocessing. We use the error-radius proxy
\begin{equation}
\begin{aligned}
b_d^{(v)}
&=
\sqrt{\frac{2u_d^{(v)}\ln(1/\delta_r)}{n_d^{(v)}}}
+\frac{3\ln(1/\delta_r)}{n_d^{(v)}},
\\
r_d^{(v)}
&=\exp(-b_d^{(v)}/\tau_r).
\end{aligned}
\label{eq:app_reliability}
\end{equation}
This score increases with effective sample size and decreases with uncertainty. The constants $\delta_r$ and $\tau_r$ are fixed using source validation.

Let $d_{ij}^{(v)}$ be a view-compatible distance, such as Jensen--Shannon distance for distributions or cosine distance for normalized semantic vectors. The reliability-adjusted edge strength is
\begin{equation}
\pi_{ij}^{(v)}
=
\operatorname{sigmoid}
\left(
-d_{ij}^{(v)}/\tau_v
+\ln(r_i^{(v)}r_j^{(v)}+\epsilon)
\right).
\label{eq:app_graph_edge}
\end{equation}
Edges below a source-validated threshold are removed. For the surviving neighborhood $\mathcal N_v(i)$, normalized aggregation gives
\begin{equation}
m_i^{(v)}
=
\frac{
h_i^{(v)}+\sum_{j\in\mathcal N_v(i)}\pi_{ij}^{(v)}W_vh_j^{(v)}
}{
1+\sum_{j\in\mathcal N_v(i)}\pi_{ij}^{(v)}
}.
\label{eq:app_graph_aggregation}
\end{equation}
The self term preserves the domain's own evidence, while reliability-weighted propagation lets low-resource domains borrow information without treating every neighbor as equally trustworthy.

\subsection{Reliability-Weighted Wasserstein Fusion}
\label{app:wasserstein}

Each propagated view representation is mapped to a probability measure $\mu_i^{(v)}$ on a shared learned support. The view weights are
\begin{equation}
\lambda_i^{(v)}
=
\frac{(r_i^{(v)})^\kappa}
{\sum_{v'}(r_i^{(v')})^\kappa}.
\label{eq:app_view_weight}
\end{equation}
The fused measure is the entropy-regularized Wasserstein barycenter~\citep{cuturi2013sinkhorn,cuturi2014fast}
\begin{equation}
\mu_i^\ast
=
\arg\min_{\mu\in\Delta}
\sum_v
\lambda_i^{(v)}
W_\varepsilon(\mu,\mu_i^{(v)};C),
\label{eq:app_barycenter}
\end{equation}
where $C$ is the ground-cost matrix induced by the shared support and $W_\varepsilon$ is the Sinkhorn-regularized transport cost. A learned readout maps $\operatorname{clr}(\mu_i^\ast)$ to $z_i\in\mathbb R^{h_z}$. Reliability affects both graph propagation and cross-view fusion, but its values are computed exclusively from source-domain observations.

\subsection{Leave-One-Domain-Out Composition}
\label{app:lodo_domain}

During a source-domain composition episode, a domain $d$ is selected as a pseudo-target. Its node and incident edges are masked before graph propagation, and $z_d$ is removed from the candidate bank. Its full source representation remains available only as a reconstruction target. The target evidence encoder constructs $\widehat z_d$ from the domain description and a simulated sparse subset of interaction inputs. Candidate retrieval and target-conditioned composition produce $\bar z_d$. The source-only training objective is
\begin{equation}
\mathcal L_{\mathrm{comp}}
=
\|\,\bar z_d-z_d\,\|_2^2
+\lambda_{\mathrm{nbr}}
\|\,s(\bar z_d,\mathcal B_{-d})-s(z_d,\mathcal B_{-d})\,\|_2^2,
\label{eq:app_composition_loss}
\end{equation}
where $s(q,\mathcal B_{-d})$ is the vector of cosine relations between $q$ and the remaining source bank. The second term prevents a reconstruction with low pointwise error but incorrect neighbors. Masking both the candidate node and its graph edges prevents trivial identity copying.

\subsection{Target Evidence and Uncertainty-Aware Retrieval}
\label{app:target_evidence}

Let $e_{\mathrm{desc}}$ be the normalized encoding of the target-domain description and let $e_{\mathrm{sup}}$ be the normalized mean encoding of the support inputs when $K>0$. Source domain $d$ has a corresponding text anchor $c_d$ and learned bank representation $z_d$. For evidence stream $v\in\{\mathrm{desc},\mathrm{sup}\}$, define
\begin{equation}
\begin{aligned}
w_d^{(v)}
&=
\frac{\exp(\kappa_v\cos(e_v,c_d))}
{\sum_{d'\in\mathcal D_s}
\exp(\kappa_v\cos(e_v,c_{d'}))},
\\
\rho_v
&=
1-\frac{-\sum_dw_d^{(v)}\ln w_d^{(v)}}
{\ln|\mathcal D_s|}.
\end{aligned}
\label{eq:app_target_stream}
\end{equation}
The reliability $\rho_v$ is high when the evidence identifies a concentrated source neighborhood and low when it is nearly uninformative. The target-evidence representation is
\begin{equation}
\widehat z_t
=
\frac{
\rho_{\mathrm{desc}}\sum_dw_d^{(\mathrm{desc})}z_d
+\ind[K>0]\rho_{\mathrm{sup}}\sum_dw_d^{(\mathrm{sup})}z_d
}{
\rho_{\mathrm{desc}}+\ind[K>0]\rho_{\mathrm{sup}}+\epsilon
}.
\label{eq:app_target_representation}
\end{equation}
Only support inputs enter Eq.~\eqref{eq:app_target_representation}; support responses do not. In zero-shot inference the description stream alone is used.

Let $r_d$ be the aggregate source reliability of candidate $d$ and let $u_t=1-\max(\rho_{\mathrm{desc}},\ind[K>0]\rho_{\mathrm{sup}})$. Retrieval confidence is
\begin{equation}
\omega_{td}
\propto
r_d\exp\!\left(
\frac{\cos(\widehat z_t,z_d)}
{\tau_0+\tau_u u_t}
\right).
\label{eq:app_retrieval_confidence}
\end{equation}
High target uncertainty increases the effective temperature and yields a less brittle candidate distribution. The top $k$ candidates form $\mathcal N_k(t)$. Their final target-conditioned contributions are those in Eq.~(8) of the main paper: retrieval determines admissible references, while attention may correct their relative weights for the current target.

\subsection{Topology-Preserving Projection}
\label{app:topology}

For a batch of $B$ composed domain representations, let $\overline Z\in\mathbb R^{B\times h_z}$ stack the centered representations and let $\overline T\in\mathbb R^{B\times h}$ stack centered mean-pooled soft-token representations. Their linear CKA~\citep{kornblith2019similarity} is
\begin{equation}
\operatorname{CKA}(\overline Z,\overline T)
=
\frac{\|\overline Z^\top\overline T\|_F^2}
{\|\overline Z^\top\overline Z\|_F
\|\overline T^\top\overline T\|_F}.
\label{eq:app_cka}
\end{equation}
For composed domains $i$ and $j$, $S^z_{ij}$ is their cosine similarity. Let $q_i$ and $k_i$ be the mean-pooled outputs obtained after the domain tokens pass through the frozen query and key projections of the language model. The symmetrized attention-interface relation is
\begin{equation}
S^A_{ij}
=
\tfrac12\left[
\cos(q_i,k_j)+\cos(q_j,k_i)
\right].
\label{eq:app_attention_relation}
\end{equation}
The topology loss is
\begin{equation}
\mathcal L_{\mathrm{topo}}
=
1-\operatorname{CKA}(\overline Z,\overline T)
+\lambda_{\mathrm{attn}}\|S^A-S^z\|_F^2.
\label{eq:app_topology_loss}
\end{equation}
The first term preserves global relational geometry across representation spaces; the second preserves relations after the frozen projections through which the model consumes the injected tokens. This is relation preservation, not a claim that the two spaces are pointwise identical or topologically homeomorphic.

\subsection{Injection and Source-Only Training}
\label{app:domain_training}

The projector uses a bottleneck MLP followed by a reshape:
\begin{equation}
T_d=\operatorname{reshape}\!\left(W_2\varphi(W_1\bar z_d)\right)
\in\mathbb R^{n_T\times h}.
\end{equation}
The tokens are inserted after the system/user-preference prefix and before the interaction query. In every source episode, the same $T_d$ conditions support and query examples and is held fixed while the task LoRA parameters adapt. The composition module and projector receive gradients from reconstruction, topology alignment, and episodic query generation. At target time they are frozen.

\section{Source-Trained User Preference Summarizer}
\label{app:summarizer}

\subsection{Leave-One-Domain-Out Training}
\label{app:summarizer_lodo}

The summarizer $f_\omega$ is trained only on source users and source domains. For a source user, one source domain is held out as a pseudo-target. The summarizer observes history from the remaining domains and generates a textual preference prompt. Held-out interactions are used to score whether the prompt improves personalization beyond the evidence from which it was constructed. This protocol trains the prompt to preserve tendencies that transfer to an unobserved domain instead of copying local topics from the input history.

\subsection{Multi-View History Perturbations}
\label{app:history_views}

Two history views for the same user are constructed by independently sampling interaction subsets, masking segments, permuting interaction order, and varying the balance among questions, responses, and metadata fields. The perturbations remove reliable access to any single surface template. Consistency is imposed on core preference statements supported by both views, while view-specific details are allowed to differ. This avoids forcing an identical prompt when the two views contain genuinely different evidence.

\subsection{Source-Only Preference Optimization}
\label{app:grpo}

The implementation uses group-relative policy optimization~\citep{shao2024deepseekmath} to train the discrete prompt generator. For one history condition $c$, sample $J$ candidate prompts $\{P_j\}_{j=1}^{J}$. Let $R_j$ be the source-only reward and define the group-normalized advantage
\begin{equation}
A_j=\frac{R_j-\overline R}{s_R+\epsilon}.
\label{eq:app_grpo_advantage}
\end{equation}
The policy objective is
\begin{equation}
\begin{aligned}
\mathcal L_{\mathrm{sum}}
&=
-\frac1J\sum_{j=1}^{J}
A_j\sum_{r=1}^{|P_j|}
\ln\pi_\omega(P_{j,r}\mid c,P_{j,<r})
\\
&\quad
+\lambda_{\mathrm{KL}}
\operatorname{KL}(\pi_\omega\Vert\pi_{\mathrm{ref}}).
\end{aligned}
\label{eq:app_grpo_loss}
\end{equation}

The reward contains four auditable source-only terms. First, a utility term measures the improvement in the log-likelihood margin between a held-out personalized response $y^+$ and a generic alternative $y^-$:
\begin{equation}
\begin{aligned}
\Delta m(P)
&=
\left[\ln p(y^+\mid x,P)-\ln p(y^-\mid x,P)\right]
\\
&\quad
-\left[\ln p(y^+\mid x)-\ln p(y^-\mid x)\right].
\end{aligned}
\label{eq:app_utility_margin}
\end{equation}
Second, a domain-leakage term rewards high entropy under a source-trained domain classifier, discouraging prompts that reveal an input domain from surface topics. Third, a cross-view stability term rewards semantic agreement between core statements extracted from two views of the same user, gated by the overlap of their evidence. Fourth, format and length terms keep the prompt concise and parseable. The total reward is
\begin{equation}
R
=
w_u r_{\mathrm{util}}
+w_d r_{\mathrm{dom}}
+w_s r_{\mathrm{stab}}
+w_f r_{\mathrm{fmt}}
+w_\ell r_{\mathrm{len}},
\label{eq:app_summary_reward}
\end{equation}
with all weights selected on source validation. The domain-leakage and stability weights are increased only after the utility and formatting rewards stabilize, which reduces collapse to generic, high-entropy summaries.

\subsection{Prompt Format and Target-Time Use}
\label{app:prompt_format}

The output is a short natural-language profile organized around repeatedly supported preferences: desired level of detail, reasoning and evidence-use tendencies, presentation and interaction preferences, and explicit negative constraints. Domain-specific entities are included only when repeated evidence supports them as a transferable preference rather than an incidental source-domain topic. The prompt contains no user identifier. If $\mathcal H_u$ is empty, a fixed neutral prompt states that no stable user preference is available. At target time the summarizer is frozen, $P_u=f_\omega(\mathcal H_u)$ is computed once, and the target support and test responses are unavailable to it.

\section{End-to-End Training and Inference}
\label{app:algorithms}

\subsection{Source-Domain Training}
\label{app:source_training_algorithm}

Algorithm~\ref{alg:app_source_training} gives the complete source-only training sequence and makes the pseudo-target masking and information boundaries explicit.

\begin{algorithm}[H]
\caption{Source-domain episodic training}
\label{alg:app_source_training}
\begin{algorithmic}[1]
\REQUIRE Source interactions, source domains $\mathcal D_s$, frozen backbone
\STATE Train the preference summarizer on source-only leave-one-domain-out prompt episodes.
\STATE Construct the three source-domain views, reliability scores, view-specific graphs, and source bank $\mathcal B$.
\FOR{each meta-training iteration}
\STATE Sample a source domain $d$ as a pseudo-target and remove its node and incident edges from the reference graph.
\STATE Construct pseudo-target evidence from its description and a sparse subset of interaction inputs.
\STATE Retrieve and compose candidates from $\mathcal B\setminus\{z_d\}$; project the composition to $T_d$.
\STATE Sample a user--domain task $\tau$ with disjoint $\mathcal S_\tau$ and $\mathcal Q_\tau$ and compute its fixed prompt $P_\tau$.
\STATE Compute $\widetilde H(\mathcal S_\tau;\theta_0)$ and $\{\widehat F_l\}$ once at $\theta_0$; stop gradients through these statistics.
\STATE Starting from $\theta_0$, take $M$ steps on Eq.~\eqref{eq:app_adapt} using $\mathcal S_\tau$.
\STATE Evaluate the adapted parameters on $\mathcal Q_\tau$ under the same $P_\tau$ and $T_d$.
\STATE Update $\theta_0$, variance parameters, graph/composition modules, and projector with the outer objective.
\ENDFOR
\STATE Freeze all shared components before target evaluation.
\end{algorithmic}
\end{algorithm}

\subsection{Unified Zero- and Few-Shot Inference}
\label{app:target_inference_algorithm}

Algorithm~\ref{alg:app_target_inference} gives the common zero- and few-shot inference path.

\begin{algorithm}[H]
\caption{Target-domain inference}
\label{alg:app_target_inference}
\begin{algorithmic}[1]
\REQUIRE Pre-target history $\mathcal H_u$, target description, query $x_t$, optional support $\mathcal S_t$, frozen shared modules
\STATE $P_u\leftarrow f_\omega(\mathcal H_u)$.
\STATE Construct $\widehat z_t$ from the target description and support inputs only.
\STATE Retrieve $\mathcal N_k(t)$ from $\mathcal B$, compose $\bar z_t$, and set $T_t\leftarrow\Pi_\psi(\bar z_t)$.
\STATE $\theta\leftarrow\theta_0$.
\IF{$K>0$}
\STATE Condition every support example on the same $[P_u;T_t]$.
\STATE Compute $\widetilde H(\mathcal S_t;\theta_0)$ and $\{\widehat F_l\}$ once and detach them.
\FOR{$m=0,\ldots,M-1$}
\STATE $\theta\leftarrow\theta-\eta_{\mathrm{test}}\nabla_\theta\mathcal L_{\mathrm{adapt}}(\theta;\mathcal S_t)$.
\ENDFOR
\ENDIF
\STATE Generate and return $G([P_u;T_t;x_t];\theta)$.
\end{algorithmic}
\end{algorithm}

\subsection{Behavior at the Two Endpoints}
\label{app:zero_few_behavior}

When $K=0$, no target gradient, entropy statistic, or Fisher statistic is computed; prediction uses $\theta_0$ with user and domain conditioning. When $K>0$, the target responses enter only the LoRA adaptation path. Increasing $K$ weakens the explicit $K^{-1}$ anchor, while the entropy and Fisher terms retain model-relative and direction-relative calibration. The method therefore implements one conditioning interface with two adaptation modes rather than two separately trained systems.

\section{Experimental Protocol}
\label{app:experiments}

\subsection{Benchmarks and Domain Splits}
\label{app:datasets}

HiCUPID~\citep{mok-etal-2025-exploring} evaluates personalized assistant responses whose relevant user attributes must be inferred from interaction history. The ten source domains are Major, Self-improvement, Social Media, Family, Forms of Living, Technology, Environment, Electronics, Religion, and Art. Politics, Music, Finance, Beauty, and Food are held out as target domains. LaMP-QA~\citep{salemi2025lamp} evaluates personalized long-form question answering using hidden user-expected aspects. The held-out target set used in the complete cross-domain analysis is Parenting, Interpersonal, Politics, Law, and Philosophy; the main paper shows the representative domains that fit within its space budget. We select these two public benchmarks because they provide complementary tests of the same claim: HiCUPID supplies controlled personalized-assistant domains and an official evaluator, whereas LaMP-QA provides naturally occurring long-form questions, diverse domain conventions, and non-unique valid answers. Their combination tests whether the method transfers across both conversational domains and task/evaluation formats rather than exploiting one benchmark's construction. The exact domain and interaction assignments are emitted by the deterministic split utility in the code archive.

For every dataset, source training, source validation, and target evaluation are separated at the domain level. Hyperparameter selection uses only source validation episodes. Within a target user--domain task, support and test interactions are disjoint. The same support indices are reused across methods and random seeds.

\begin{figure*}[t]
\centering
\begin{subfigure}[b]{0.47\textwidth}
\centering
\includegraphics[width=\linewidth]{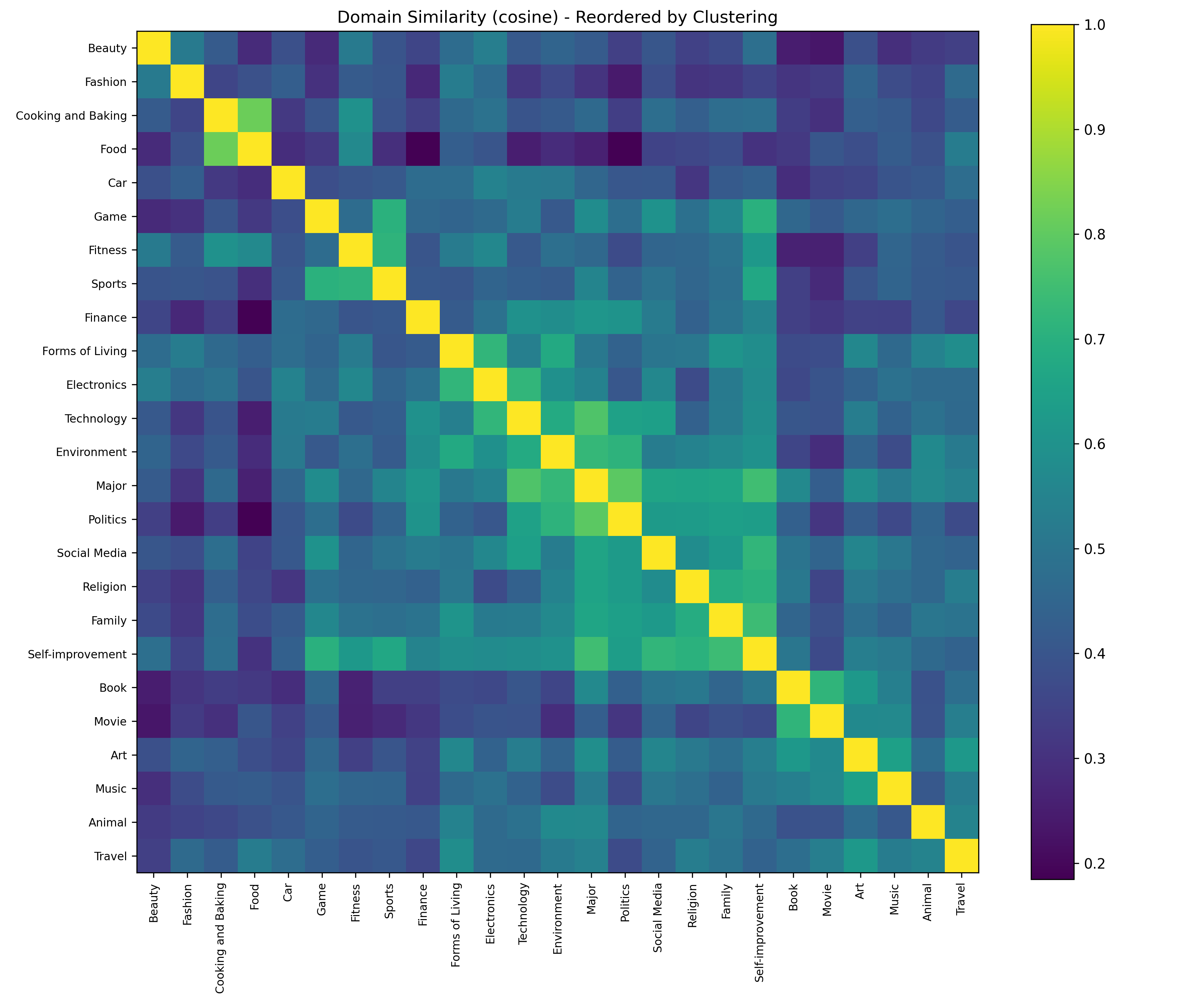}
\caption{HiCUPID.}
\end{subfigure}
\hfill
\begin{subfigure}[b]{0.47\textwidth}
\centering
\includegraphics[width=\linewidth]{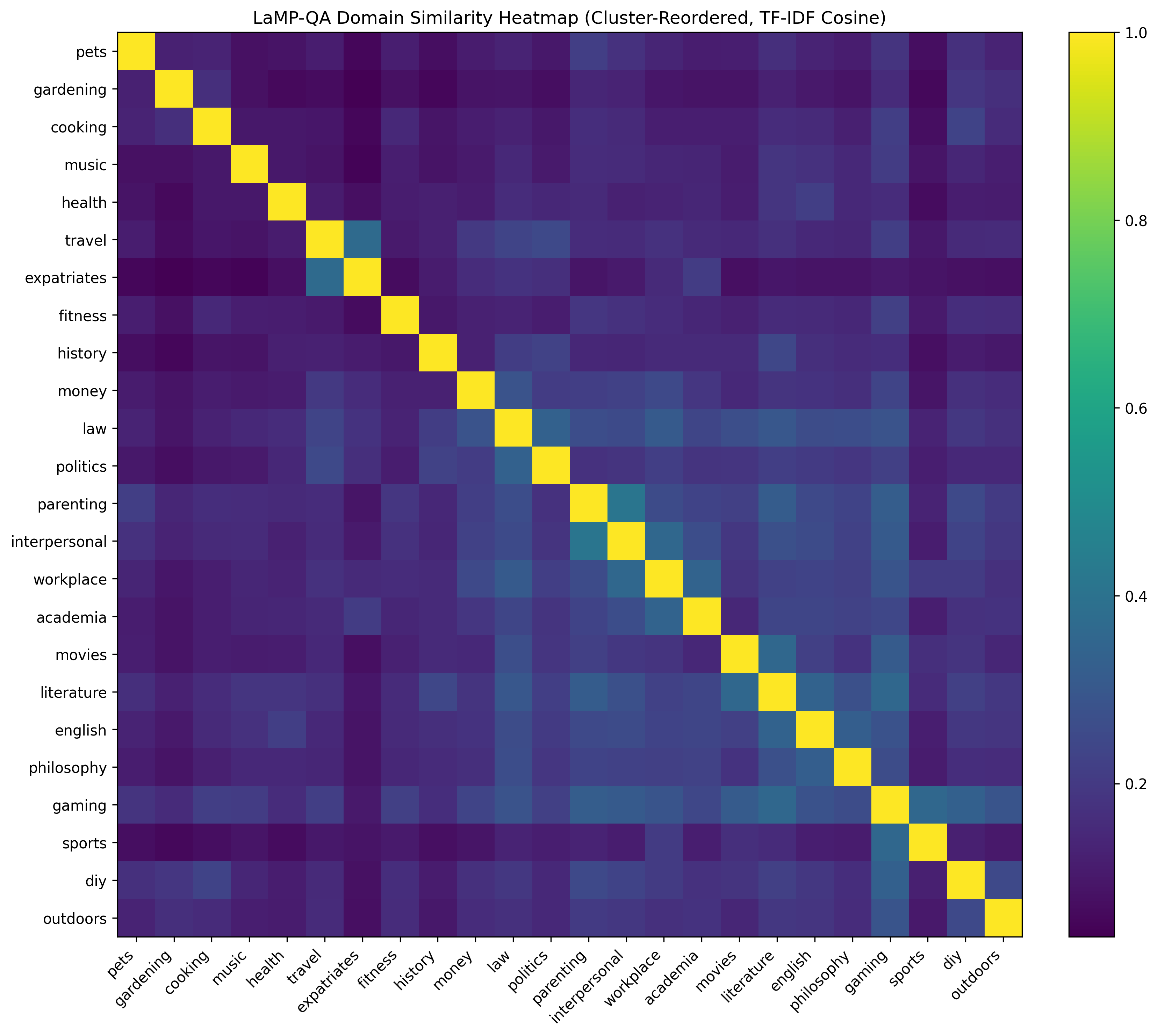}
\caption{LaMP-QA.}
\end{subfigure}
\caption{Inter-domain similarity audit for the two benchmark partitions. Darker cells indicate larger cosine similarity between aggregate held-out interaction representations.}
\label{fig:app_domain_heatmaps}
\end{figure*}

Figure~\ref{fig:app_domain_heatmaps} visualizes the aggregate inter-domain similarities used only to audit the chosen source/target partition. The target domains span both near and distant relations to the source set, so the evaluation is not restricted to trivial semantic neighbors. These visualizations are descriptive; they are not supplied to any compared method.

\subsection{Baseline Implementations}
\label{app:baselines}

\begin{itemize}
\item \textbf{SFT/LoRA} adapts the source-trained model on the target support set without episodic initialization.
\item \textbf{LaMP}~\citep{salemi-etal-2024-lamp} retrieves user-history evidence and prepends it to the generator input.
\item \textbf{GPG}~\citep{zhang2024guided} generates a textual profile from user history and conditions generation on that profile.
\item \textbf{SPT}~\citep{huang2024selective,lester-etal-2021-power} learns continuous prompt vectors while keeping the backbone frozen.
\item \textbf{DEP}~\citep{qiu-etal-2025-latent} uses a latent preference representation for continuous conditioning.
\item \textbf{IDL}~\citep{cheng2024dialogues} composes shared low-rank components for user-level adaptation.
\item \textbf{OPPU}~\citep{tan-etal-2024-democratizing} optimizes a user-specific LoRA module with its original objective on exactly the same target support set. Its source-trained backbone is frozen.
\item \textbf{PROPER}~\citep{zhang2025proper} performs its published progressive/group-level parameter adaptation using the same support examples.
\item \textbf{CoPL}~\citep{choi-etal-2025-copl} applies its published contrastive personalization objective under the same source/target split.
\item \textbf{BiPO}~\citep{NEURIPS2024_58cbe393} performs bidirectional preference steering at inference.
\item \textbf{CHAMELEON}~\citep{zhang2025personalize} edits internal representations using its published personalization procedure.
\item \textbf{OPAD}~\citep{zhu2025fly} performs on-the-fly preference alignment with its original inference objective.
\item \textbf{Proactive}~\citep{zhang2025towards} uses proactive clarification or preference elicitation before response generation.
\item \textbf{MAML}~\citep{pmlr-v70-finn17a} is Model-Agnostic Meta-Learning over the same LoRA parameter subset used by the other parameter-adaptation methods.
\item \textbf{Vanilla Meta-LoRA} uses the same source episodes, LoRA parameterization, number of inner steps, and conditioning inputs as \method, but its inner loop minimizes only support NLL; it has no $K^{-1}$ anchor, entropy calibration, learned depth-group variances, or Fisher weighting.
\end{itemize}

Each baseline uses its original objective and receives no target interaction unavailable to another method in the same shot setting. Decoding parameters, maximum context, support partitions, test instances, and pairwise evaluation order are shared. When a method cannot natively use a component of our framework, that component is not added unless the experiment explicitly reports a controlled plug-in analysis.

\begin{table*}[t]
\centering
\small
\setlength{\tabcolsep}{4.2pt}
\renewcommand{\arraystretch}{1.08}
\begin{tabular}{p{0.17\textwidth}p{0.76\textwidth}}
\toprule
Method & Final method-specific setting \\
\midrule
SFT/LoRA & LoRA rank/scaling/dropout $=16/32/0.05$ on \texttt{q\_proj}, \texttt{v\_proj}; five target steps; learning rate $5\times10^{-3}$. \\
LaMP & BM25 retrieval with the top five history items, chronological tie breaking, and no target-parameter update. \\
GPG & Source-trained profile generator; maximum profile length $300$ words; one fixed profile per target episode and no target-parameter update. \\
SPT & Eight trainable prompt tokens; five target steps; learning rate $5\times10^{-3}$; backbone frozen. \\
DEP & Latent dimension $256$; AdamW learning rate $10^{-5}$; five source-training epochs; published latent-conditioning objective. \\
IDL & Four shared low-rank components of rank $16$; source-learned composition weights; five target steps at $5\times10^{-3}$. \\
OPPU & User LoRA rank/scaling/dropout $=16/32/0.05$; five updates at $5\times10^{-3}$ on the target support set using the original OPPU loss. \\
CoPL & Graph stage: $30$ epochs, learning rate $10^{-4}$, regularization weight $10^{-6}$; LoRA reward stage: rank $8$, scaling $16$, dropout $0.1$, learning rate $10^{-5}$, maximum length $512$. \\
PROPER / GMoE & Stage 1: rank/scaling $=8/16$, learning rate $3\times10^{-4}$, two epochs; Stage 2: $8/16$, $2\times10^{-4}$, three epochs; Stage 3: $4/8$, dropout $0.05$, $2\times10^{-4}$, one epoch. \\
BiPO / CHAMELEON & Released inference objectives and source-selected default control strengths; no target gradient update; the common deterministic decoding configuration is used. \\
OPAD / Proactive & Released online-control objectives with at most the same $K$ target interactions; source-selected default control coefficients and the common decoding configuration. \\
MAML & Five inner steps at $10^{-4}$ and outer learning rate $5\times10^{-5}$; the same LoRA parameter subset and source episodes as Meta-LoRA. \\
Vanilla Meta-LoRA & LoRA rank/scaling/dropout $=16/32/0.05$; five inner steps at $10^{-4}$ and outer learning rate $5\times10^{-5}$; no anchoring term. \\
\bottomrule
\end{tabular}
\caption{Final baseline settings. Entries list deviations from the shared backbone, data splits, context limit, and decoding protocol in Table~\ref{tab:hyperparameters}; unlisted options retain the cited implementation's defaults.}
\label{tab:baseline_hyperparameters}
\end{table*}

\subsection{Metrics}
\label{app:metrics}

\paragraph{Win Rate.}
For test instance $i$, a fixed pairwise evaluator compares the generated personalized response with the corresponding benchmark reference under randomized response order. Pairwise, position-randomized LLM evaluation is preferred to an unblinded absolute rating because it reduces scale-calibration and presentation biases~\citep{zheng2023judging}. Let $w_i=1$ for a win, $w_i=0$ for a loss, and $w_i=1/2$ for a persistent tie under the fixed tie policy. Then
\begin{equation}
\mathrm{WR}=100\,\frac1N\sum_{i=1}^{N}w_i.
\label{eq:app_wr}
\end{equation}
HiCUPID uses its official evaluator. LaMP-QA uses the fixed hidden-aspect pairwise rubric described below.

\paragraph{Personal Gap.}
Let $e(\cdot)$ be a frozen Sentence-BERT encoder~\citep{reimers2019sentencebert}, $y_i^+$ the personalized target, $y_i^0$ the generic alternative, and $\widehat y_i$ the generated response. The judge-independent specificity margin is
\begin{equation}
\mathrm{PG}
=
\frac1N\sum_{i=1}^{N}
\left[
\cos(e(\widehat y_i),e(y_i^+))
-\cos(e(\widehat y_i),e(y_i^0))
\right].
\label{eq:app_pg}
\end{equation}
A positive value means that the response is closer to the personalized target than to a generic alternative.

\paragraph{Fine-grained quality and CCS.}
A fixed method-blinded rubric, following the multidimensional evaluation practice of HiCUPID~\citep{mok-etal-2025-exploring}, scores Utility $U_i$, Honesty $H_i$, and Personal Fit $P_i$ on $\{0,1,2\}$. The Calibrated Composite Score is
\begin{equation}
\mathrm{CCS}
=
\frac1N\sum_{i=1}^{N}
\left(0.3U_i+0.3H_i+0.4P_i\right).
\label{eq:app_ccs}
\end{equation}
The larger weight on Personal Fit reflects the personalization objective while retaining explicit utility and factuality terms. The same prompt, weights, and judge checkpoint are fixed before comparing methods.

\paragraph{Cross-domain degradation.}
If $\mathrm{WR}_{\mathrm{src}}$ is the mean source-domain WR and $\overline{\mathrm{WR}}_{\mathrm{tgt}}$ is the unweighted mean over held-out target domains, then
\begin{equation}
\Delta\mathrm{WR}
=
\mathrm{WR}_{\mathrm{src}}
-\overline{\mathrm{WR}}_{\mathrm{tgt}}.
\label{eq:app_delta_wr}
\end{equation}
All arithmetic is performed from unrounded domain-level values; displayed values are rounded only after aggregation.

\subsection{Evaluator and Human-Agreement Protocol}
\label{app:evaluators}

The HiCUPID WR calculation follows its official evaluator~\citep{mok-etal-2025-exploring}. LaMP-QA lacks a single gold answer, so evaluation follows its hidden-aspect protocol~\citep{salemi2025lamp} and asks the frozen evaluator to compare two responses against the annotated expected aspects. Method names are removed and answer order is randomized. PG is computed without an LLM judge.

For the reliability study, 100 test instances are sampled independently from each benchmark. Human assessors receive the same query, permissible user evidence, and rubric but no method identity. Scores from the fixed Qwen3-32B evaluator~\citep{qwen3technicalreport} are correlated with human scores and an independent GPT-4o evaluation using Pearson's $r$. The resulting agreement is reported in Appendix~\ref{app:evaluator_agreement}; it evaluates the measurement procedure rather than the superiority of any judge model.

\subsection{Implementation and Hyperparameters}
\label{app:hyperparameters}

Table~\ref{tab:hyperparameters} records the final shared settings, and Table~\ref{tab:hyperparameter_search} reports the complete grid for every tunable coefficient, capacity, and learning rate in the proposed modules. Architecture-imposed quantities, deterministic decoding, hardware, and cited baseline defaults are fixed rather than optimized.

Grid search uses nested source-only validation. Source domains are rotated as pseudo-targets; each is removed from the reference bank, and its query examples remain disjoint from support. Every candidate in a stage receives identical episode manifests, training budget, and seeds. We select common LoRA/meta-optimization settings first, followed by the evidence-calibrated objective, domain retrieval and reliability, soft-token projection, and summarizer optimization. The primary criterion is equal-domain-weighted mean WR across source-validation pseudo-targets. If candidates are within $0.2$ WR points, CCS is the tie-breaker; a remaining tie favors the smaller model or weaker regularizer. The selected value is frozen before the next stage. The complete configuration is then retrained on source data and used unchanged for every held-out target domain and shot setting. Target labels and target-domain development scores never enter search, early stopping, or checkpoint choice.

\begin{table*}[t]
\centering
\small
\setlength{\tabcolsep}{4.2pt}
\renewcommand{\arraystretch}{1.07}
\begin{tabular}{p{0.18\textwidth}p{0.25\textwidth}p{0.50\textwidth}}
\toprule
Component & Setting & Value or rule \\
\midrule
Backbone & Default / cross-model & LLaMA-3-8B-Instruct / Qwen3-8B~\citep{qwen3technicalreport}; backbone weights frozen \\
Numerics & Precision / clipping & bfloat16 forward-backward; trainable LoRA states in FP32; gradient norm clipped to $1.0$ \\
Optimization & Optimizer & AdamW, $\beta_1=0.9$, $\beta_2=0.999$ \\
Optimization & Schedule & cosine decay; weight decay $0.005$; warmup ratio $0.05$; early-stopping patience $3$ \\
Optimization & Epochs / effective batch & at most $20$ / $64$ \\
Optimization & Maximum sequence length & $2048$ \\
Generation & Decoding & greedy (\texttt{do\_sample=False}); maximum $512$ new tokens; batch size $1$ \\
LoRA & Rank / scaling / dropout & $16$ / $32$ / $0.05$ \\
LoRA & Target modules & \texttt{q\_proj}, \texttt{v\_proj} \\
Meta-learning & Inner steps / inner LR & $M=5$ / $10^{-4}$ \\
Meta-learning & Outer LR / target LR & $5\times10^{-5}$ / $5\times10^{-3}$ \\
Meta-learning & Training support sizes & sampled from $\{1,3,5,10\}$ \\
PAC-Bayes surrogate & Analysis map / entropy & $c=5.0$ in Eq.~\eqref{eq:app_bounded_loss}; $\alpha=1.0$; Fisher floor $10^{-8}$ \\
PAC-Bayes surrogate & Depth groups / variance & layers $1$--$8$, $9$--$24$, $25$--$32$; variance bounds $[0.001,0.25]$; initialization $\{0.05,0.15,0.08\}$ \\
Domain bank & Representation / retrieval & $h_z=256$; candidate cap $k=5$; edge threshold $0.65$; reliability temperature $0.10$; retrieval-prior coefficient $\eta=0.5$ \\
Projector & Capacity / topology weights & one domain soft token; bottleneck $128$; $\lambda_{\mathrm{attn}}=0.5$; outer topology weight $0.1$ \\
Summarizer & GRPO optimization & $4$ candidates; $6$ epochs; learning rate $10^{-5}$; KL coefficient $0.04$; maximum $400$ new tokens \\
Summarizer & Reward weights & utility/domain/stability/format/length $=1.0/0.5/0.5/0.5/0.1$ \\
Evaluation & Runs / seeds & $10$ runs; seeds $42$--$51$ \\
Evaluation & Fine-grained evaluator & Qwen3-32B; one frozen checkpoint, prompt, decoding rule, and answer-order policy for all methods \\
Hardware & Accelerators / host & $2\times$ NVIDIA A100 80GB; Intel Xeon Silver 4210 CPU; $256$ GB host RAM \\
Software & System / libraries & Ubuntu 20.04; CUDA 11.8; Python 3.10; PyTorch 2.1.0; Transformers 4.51.3; PEFT 0.15.2; Accelerate 1.6.0 \\
\bottomrule
\end{tabular}
\caption{Final implementation settings. Values governing the new modules are selected exclusively on source-validation episodes.}
\label{tab:hyperparameters}
\end{table*}

\begin{table*}[t]
\centering
\small
\setlength{\tabcolsep}{4.0pt}
\renewcommand{\arraystretch}{1.06}
\begin{tabular}{p{0.22\textwidth}p{0.50\textwidth}p{0.18\textwidth}}
\toprule
Parameter & Values tried & Selected \\
\midrule
LoRA rank / dropout & $\{8,16,32\}$ / $\{0,0.05,0.10\}$ & $16$ / $0.05$ \\
Inner steps / learning rate & $\{1,3,5,8\}$ / $\{5{\times}10^{-5},10^{-4},5{\times}10^{-4}\}$ & $5$ / $10^{-4}$ \\
Outer / target learning rate & $\{10^{-5},5{\times}10^{-5},10^{-4}\}$ / $\{10^{-3},5{\times}10^{-3},10^{-2}\}$ & $5{\times}10^{-5}$ / $5{\times}10^{-3}$ \\
Entropy coefficient $\alpha$ & $\{0,0.5,1,2\}$ & $1$ \\
Variance initialization & $\{(.05,.15,.08),(.08,.12,.10),(.10,.10,.10),(.15,.05,.12)\}$ & $(.05,.15,.08)$ \\
Candidate cap $k$ & $\{1,3,5,8\}$ & $5$ \\
Graph edge threshold & $\{0.50,0.55,0.60,0.65,0.70\}$ & $0.65$ \\
Reliability temperature & $\{0.05,0.10,0.20\}$ & $0.10$ \\
Retrieval-prior coefficient $\eta$ & $\{0,0.5,1.0\}$ & $0.5$ \\
Soft-token count & $\{1,4,8\}$ & $1$ \\
Projector bottleneck & $\{64,128,256\}$ & $128$ \\
Attention-alignment weight & $\{0.1,0.5,1.0\}$ & $0.5$ \\
GRPO group size / KL coefficient & $\{2,4,8\}$ / $\{0.01,0.04,0.10\}$ & $4$ / $0.04$ \\
\bottomrule
\end{tabular}
\caption{Development search spaces and selected values. Each brace-delimited set is the complete set tried for that parameter; fixed choices are reported in Table~\ref{tab:hyperparameters}.}
\label{tab:hyperparameter_search}
\end{table*}

\paragraph{Randomness control.}
For run seed $s\in\{42,\ldots,51\}$, the same value initializes Python's \texttt{random}, NumPy, PyTorch CPU, and every CUDA device, and is passed to episode sampling, support-set construction, data-loader shuffling, and model initialization. We set \texttt{PYTHONHASHSEED=$s$}, disable cuDNN benchmarking, enable deterministic cuDNN kernels, and request deterministic PyTorch algorithms with warning mode for operations without a deterministic CUDA implementation. Data-loader workers use seed $s+\texttt{worker\_id}$, and the generator supplied to each loader is seeded by $s$. Split manifests and nested support indices are generated once and reused by all methods. Greedy decoding removes sampling randomness at generation time; the response-order permutation used by the pairwise evaluator is separately stored for every instance and seed.

\subsection{Statistical Reporting and Fairness Controls}
\label{app:statistics}

WR is reported as mean and standard deviation over ten independent seeds. Every method is evaluated on identical target support/test indices. CCS and PG are first computed per test instance and then aggregated using the same indices. Paired comparisons use a two-sided Wilcoxon signed-rank test on matched seed-level scores and a $10{,}000$-resample paired bootstrap interval over matched test instances; Holm correction is applied when one analysis covers multiple domains. Relative improvements use the strongest competing method in the same cell as the denominator. Ablations that alter source training are retrained from scratch rather than disabled only at inference. The code archive implements paired tests, bootstrap intervals, deterministic response ordering, and nested support sampling.

\section{Extended Analyses}
\label{app:analysis}

\subsection{Exact Ablation Definitions}
\label{app:ablation_definitions}

\begin{itemize}
\item \textbf{Textual conditioning only} retains $P_u$ but removes domain soft tokens and target parameter adaptation.
\item \textbf{+ domain soft tokens} adds the complete graph-guided domain-prior path but still uses $\theta_0$ without target LoRA updates.
\item \textbf{+ vanilla Meta-LoRA} adds standard support-loss-only episodic LoRA adaptation.
\item \textbf{Full model} replaces the vanilla inner objective with Eq.~\eqref{eq:app_adapt}.
\item \textbf{Without \(K^{-1}\) scaling} replaces $1/K$ by a source-validated constant while retaining entropy and Fisher terms.
\item \textbf{Without entropy calibration} sets $\alpha=0$ but retains $1/K$.
\item \textbf{Shared group variance} replaces all learned $\sigma_g^2$ values by one source-learned scalar.
\item \textbf{Without Fisher weighting} replaces $\widehat F_l$ by the identity matrix, reducing the anchor to a depth-weighted Euclidean distance.
\item \textbf{Without graph reliability} removes reliability from edge construction, aggregation, and barycenter weighting while preserving the same view features.
\item \textbf{Without topology preservation} sets the CKA and attention-alignment weights to zero while retaining the projector architecture and token count.
\end{itemize}

These definitions separate cumulative contribution from component ablation. In the main-paper cumulative block, domain tokens raise zero-shot WR from $55.8$ to $59.0$; in the 5-shot regime, adding domain tokens, vanilla Meta-LoRA, and the complete evidence-calibrated objective raises WR from $59.3$ to $63.4$, $68.4$, and $73.2$, respectively. Thus the gains cannot be attributed only to a stronger text-conditioned base model. Because a changed inner objective also changes the learned $\theta_0$, zero-shot performance may change even though zero-shot inference itself performs no target update.

\subsection{Cross-Backbone Generalization on Qwen3}
\label{app:qwen3_results}

Table~\ref{tab:app_qwen3} reports the complete cross-model experiment underlying the Qwen3 result summarized in the main paper. Every method uses the same Qwen3-8B checkpoint~\citep{qwen3technicalreport}, target support examples, greedy decoding configuration, and evaluation instances; backbone-dependent adapters and the soft-token projector are reinitialized and source-trained rather than copied from the LLaMA experiment. The complete method attains $70.9$ WR and the smallest cross-domain degradation, while improving all three fine-grained dimensions. This matched replacement isolates portability across model families from differences in data or evaluation.

\begin{table*}[t]
\centering
\small
\setlength{\tabcolsep}{4.2pt}
\renewcommand{\arraystretch}{1.08}
\begin{tabular}{lcccccc}
\toprule
Method & WR $\uparrow$ & PG $\uparrow$ & Utility $\uparrow$ & Honesty $\uparrow$ & Personal Fit $\uparrow$ & $\Delta$WR $\downarrow$ \\
\midrule
LoRA & $24.8{\pm}1.4$ & .004 & .92 & 1.02 & .70 & 24.5 \\
DEP & $50.7{\pm}1.2$ & .033 & 1.04 & 1.15 & .98 & 17.6 \\
OPPU & $47.1{\pm}1.3$ & .039 & 1.05 & 1.16 & .93 & 22.3 \\
PROPER & $58.5{\pm}1.1$ & .051 & 1.32 & 1.27 & 1.09 & 19.2 \\
\midrule
Ours & $\mathbf{70.9{\pm}0.8}$ & \textbf{.073} & \textbf{1.56} & \textbf{1.52} & \textbf{1.33} & \textbf{6.7} \\
\bottomrule
\end{tabular}
\caption{Cross-domain personalization on the Qwen3-8B backbone. All values are from the matched Qwen3 experiment; WR is mean $\pm$ standard deviation over ten seeds.}
\label{tab:app_qwen3}
\end{table*}

\subsection{Per-Domain Personalization Specificity}
\label{app:pg_breakdown}

Table~\ref{tab:app_hicupid_pg} expands the HiCUPID Personal Gap result by domain. The complete method has the largest judge-independent specificity margin in every target domain; its target-domain mean is $0.091$, matching the 5-shot aggregate in the main-paper ablation table after rounding. The particularly large Music and Beauty margins show that the WR gain is accompanied by stronger distinction from generic answers rather than only improved fluency.

\begin{table*}[t]
\centering
\small
\setlength{\tabcolsep}{4.2pt}
\renewcommand{\arraystretch}{1.08}
\begin{tabular}{lcccccc}
\toprule
Method & Avg. source & Politics & Music & Finance & Beauty & Food \\
\midrule
SFT & .054 & .009 & .009 & .004 & .007 & .002 \\
LaMP & .066 & .024 & .016 & .009 & .022 & .012 \\
GPG & .069 & .041 & .035 & .027 & .036 & .032 \\
SPT & .062 & .031 & .028 & .025 & .026 & .023 \\
IDL & .064 & .043 & .041 & .035 & .032 & .033 \\
OPPU & .079 & .059 & .034 & .033 & .037 & .032 \\
DEP & .078 & .030 & .023 & .039 & .039 & .027 \\
CoPL & .081 & .061 & .017 & .029 & .057 & .025 \\
PROPER & .085 & .062 & .059 & .049 & .046 & .069 \\
Ours & \textbf{.093} & \textbf{.081} & \textbf{.113} & \textbf{.071} & \textbf{.093} & \textbf{.098} \\
\bottomrule
\end{tabular}
\caption{HiCUPID 5-shot Personal Gap by domain. Higher values indicate greater semantic proximity to the personalized target relative to the generic alternative.}
\label{tab:app_hicupid_pg}
\end{table*}

\subsection{Sample Efficiency and Parameter Displacement}
\label{app:sample_efficiency}

Sample efficiency is evaluated at $K\in\{0,1,3,5,10\}$ using nested support sets: the $K=1$ example is contained in the $K=3$ set, and so forth. This prevents differences between shot levels from being driven by unrelated sampled examples. Alongside WR, we measure normalized parameter displacement
\begin{equation}
D_K
=
\frac{
\left(\sum_l\|\theta_{K,l}^{\ast}-\theta_{0,l}\|_2^2\right)^{1/2}
}{
\left(\sum_l\|\theta_{0,l}\|_2^2\right)^{1/2}+\epsilon
}.
\label{eq:app_displacement}
\end{equation}
The intended behavior is not uniformly small $D_K$. We test whether the controlled objective yields smaller displacement than vanilla Meta-LoRA when evidence is sparse or uncertain and whether this gap narrows as support evidence increases.

\begin{figure}[t]
\centering
\includegraphics[width=\columnwidth]{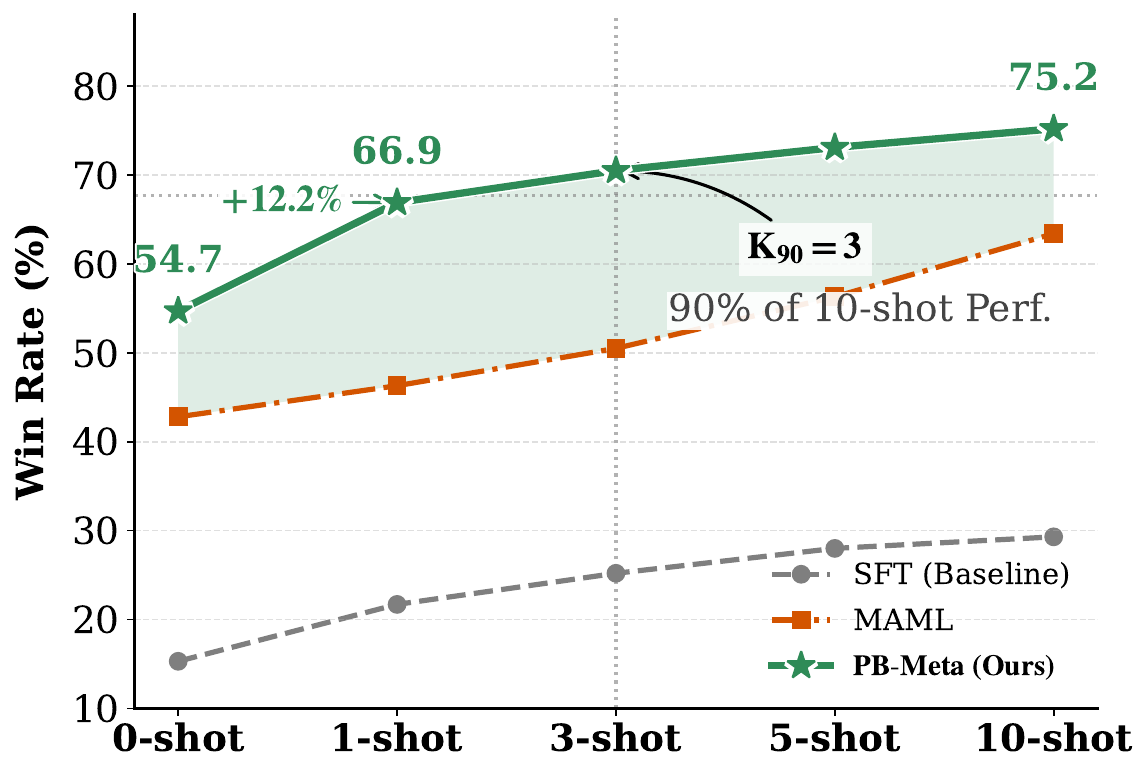}
\caption{Win Rate across $K\in\{0,1,3,5,10\}$. Every larger support set contains the smaller one for the same user, domain, and seed.}
\label{fig:app_sample_efficiency}
\end{figure}

Figure~\ref{fig:app_sample_efficiency} reports the performance trajectory used in the main paper. The proposed method increases from $54.7$ WR at $K=0$ to $66.9$ with one interaction and reaches approximately $90\%$ of its 10-shot performance by $K=3$. The nested-support construction makes each increase attributable to additional evidence rather than a different draw of support examples.

\subsection{Optimization Stability}
\label{app:optimization_stability}

\begin{figure}[t]
\centering
\includegraphics[width=\columnwidth]{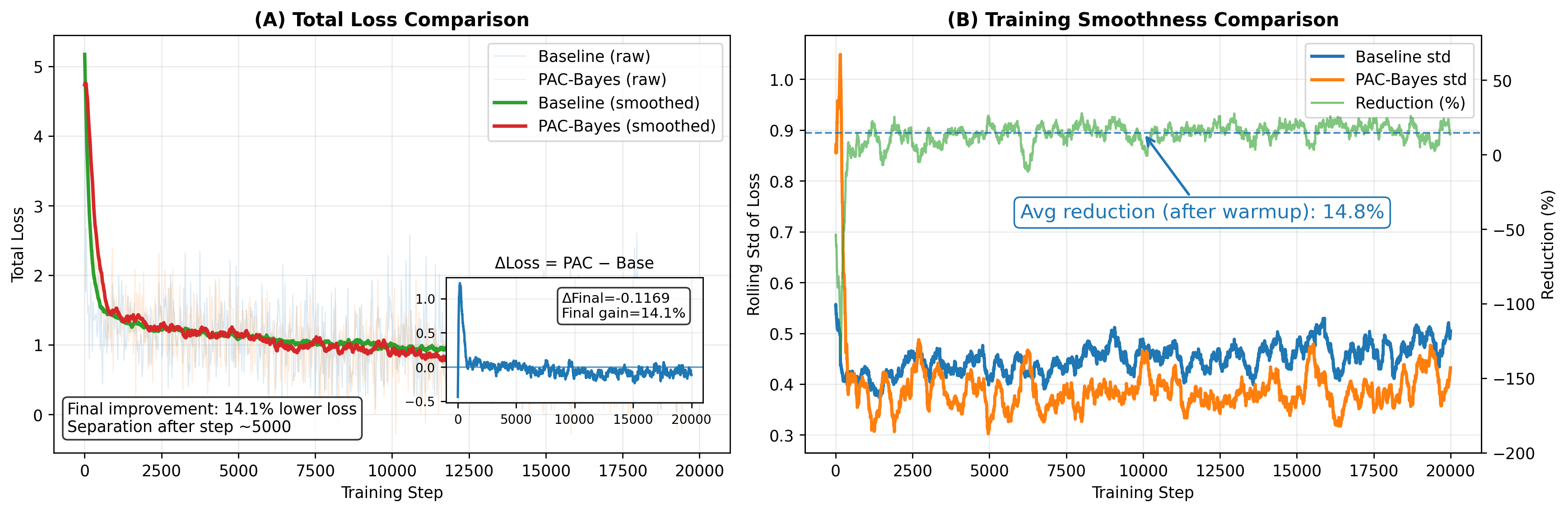}
\caption{Matched source-training dynamics. (A) Total loss. (B) Rolling loss standard deviation. The vertical marker denotes the end of warmup.}
\label{fig:app_pac_dynamics}
\end{figure}

Figure~\ref{fig:app_pac_dynamics} compares the source meta-training trajectories of vanilla Meta-LoRA and the complete evidence-calibrated objective under identical batches and optimization settings. After warmup, the complete objective attains a $14.1\%$ lower terminal loss and a $14.8\%$ lower rolling loss standard deviation. This is an optimization diagnostic, not a separate generalization guarantee: it shows that the structured anchor changes the observed training dynamics in the intended direction, while the target-domain ablations in the main paper establish its predictive contribution.

\subsection{Topology Metrics}
\label{app:topology_metrics}

Three metrics test complementary properties of the projection. CKA measures global relation alignment. Neighborhood Recall@$k$ is
\begin{equation}
\operatorname{NR@}k
=
\frac1{|\mathcal D|}
\sum_{d\in\mathcal D}
\frac{
|\mathcal N_k^z(d)\cap\mathcal N_k^T(d)|
}{k},
\label{eq:app_neighbor_recall}
\end{equation}
where the two neighborhoods are constructed in the domain and projected-token spaces. Attention Relation Error is the normalized Frobenius error $\|S^A-S^z\|_F^2/|\mathcal D|^2$. Reporting all three prevents a high global alignment score from obscuring poor local retrieval structure or poor fidelity at the actual attention interface.

\begin{table}[t]
\centering
\small
\setlength{\tabcolsep}{4.0pt}
\begin{tabular}{lcc}
\toprule
Projection objective & Attn.\ error $\downarrow$ & 0-shot WR $\uparrow$ \\
\midrule
No topology constraint & 0.31 & 57.6 \\
CKA + attention alignment & \textbf{0.07} & \textbf{59.5} \\
\bottomrule
\end{tabular}
\caption{Endpoint effect of topology preservation under otherwise identical zero-shot settings.}
\label{tab:app_topology_endpoints}
\end{table}

Component-wise analysis shows complementary effects: CKA alignment primarily improves global relational consistency, whereas attention-space alignment more directly reduces error after the tokens pass through the frozen query and key projections. Combining both yields the highest CKA and Neighborhood Recall@$5$ among the tested variants, reduces attention relation error from $0.31$ to $0.07$, and improves zero-shot WR from $57.6$ to $59.5$. The generation gain indicates that relation preservation is functionally useful rather than only improving geometric agreement.

\begin{figure*}[t]
\centering
\begin{subfigure}[b]{0.47\textwidth}
\centering
\includegraphics[width=\linewidth]{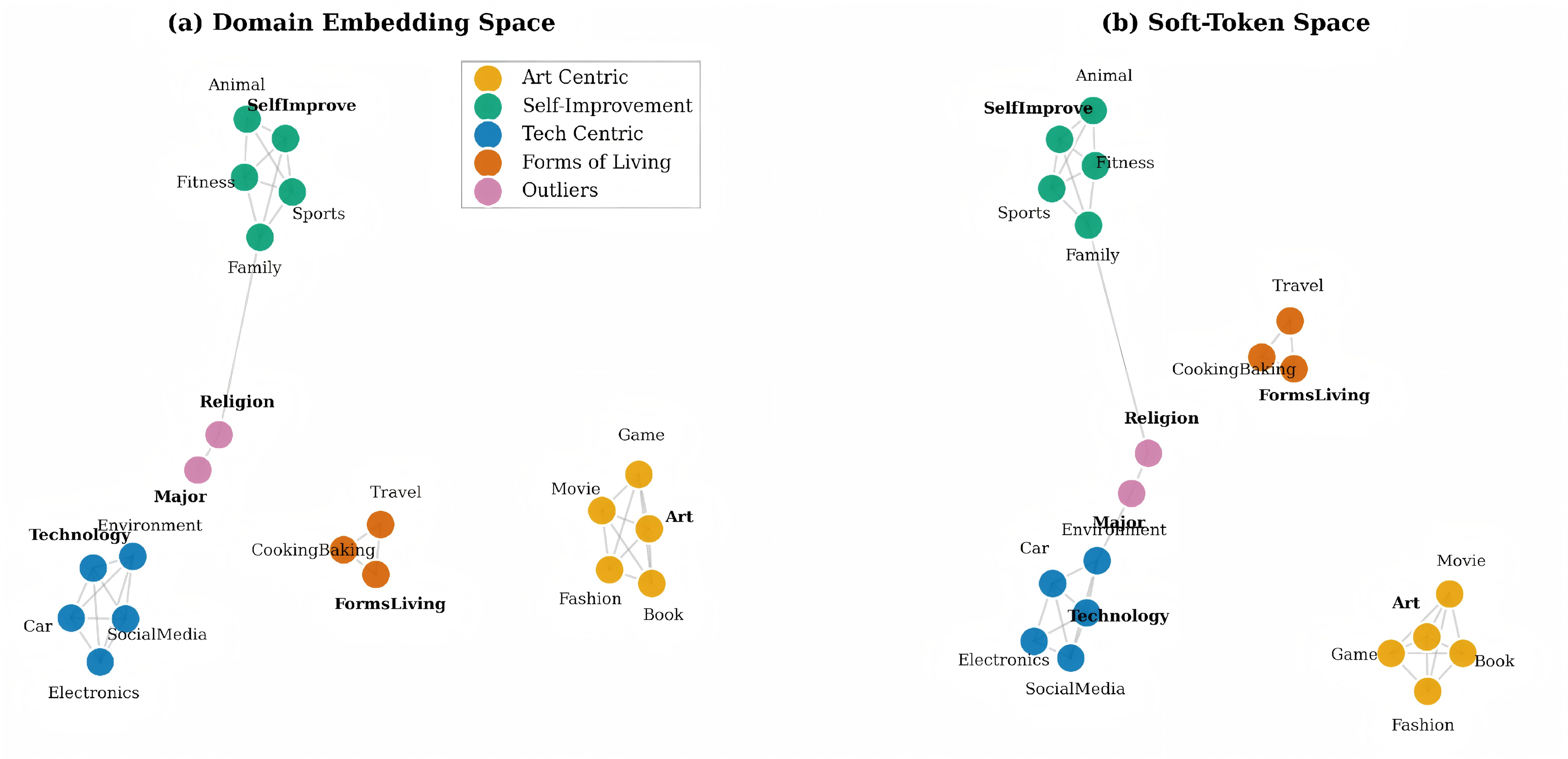}
\caption{Two-dimensional layouts and mutual-neighbor edges.}
\end{subfigure}
\hfill
\begin{subfigure}[b]{0.47\textwidth}
\centering
\includegraphics[width=\linewidth]{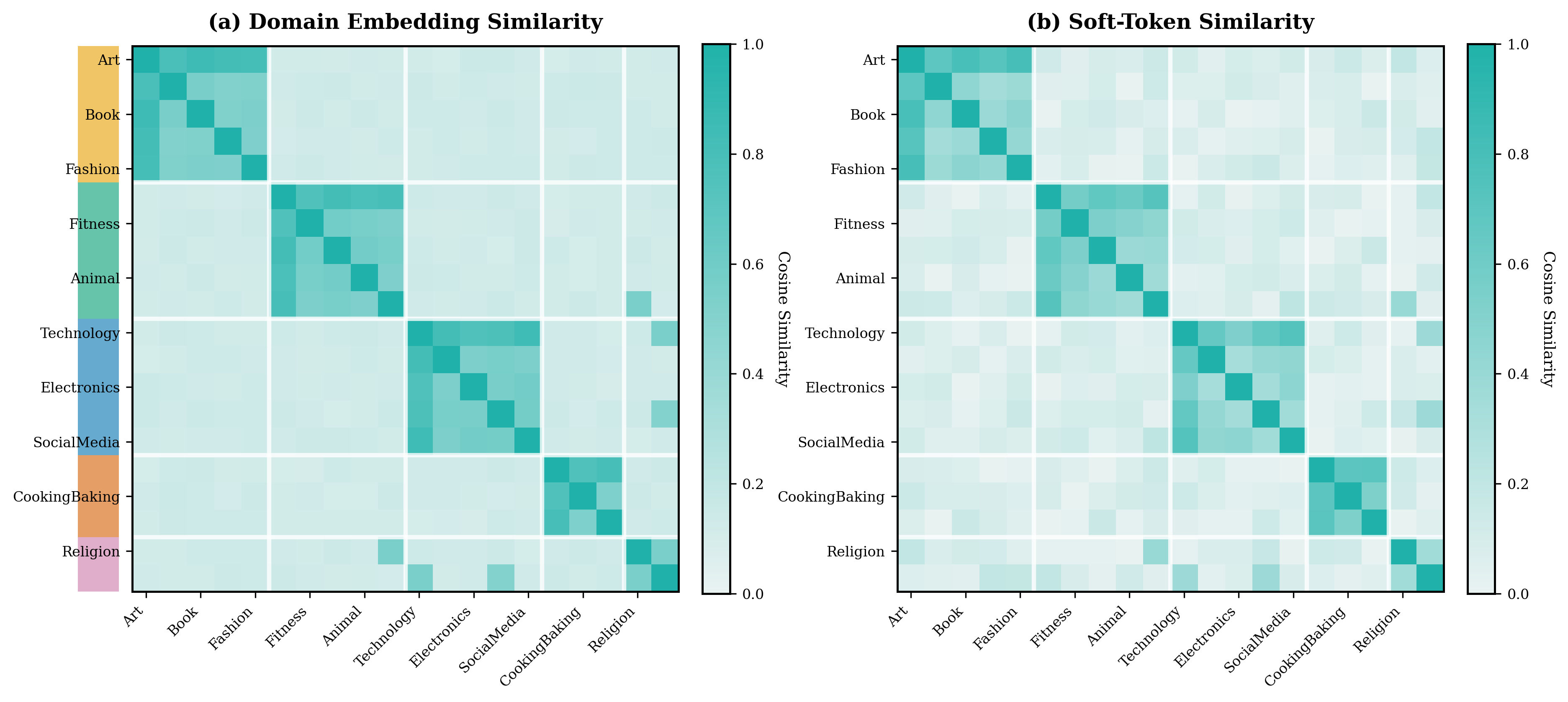}
\caption{Domain--domain cosine-similarity matrices.}
\end{subfigure}
\caption{Relations before and after topology-preserving soft-token projection. The analysis complements CKA, Neighborhood Recall@$5$, and attention-relation error with a direct visual comparison.}
\label{fig:app_topology_visual}
\end{figure*}

Figure~\ref{fig:app_topology_visual} provides the corresponding qualitative audit. Neighborhood connectivity and block structure are retained after projection; localized deviations reflect the projective bottleneck and generation supervision rather than a pointwise identity map.

\subsection{Attention Attribution to Domain Soft Tokens}
\label{app:attention_attribution}

\begin{figure}[t]
\centering
\includegraphics[width=\columnwidth]{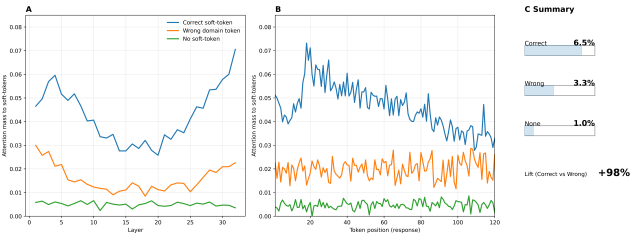}
\caption{Attention assigned to correct, wrong-domain, and non-informative soft-token conditions across layers, decoding positions, and the aggregate comparison.}
\label{fig:app_soft_token_attention}
\end{figure}

To test whether the injected token is actually read, let $\mathcal S_T$ be the soft-token positions and $\mathcal A$ all available context positions. At layer $l$ and decoding step $t$, we measure $a_{l,t}=\sum_{j\in\mathcal S_T}\operatorname{Attn}_{l,t\rightarrow j}/\sum_{j\in\mathcal A}\operatorname{Attn}_{l,t\rightarrow j}$. We compare the correct target-domain token, a token from the least similar source domain, and an equal-length non-informative placeholder. As shown in Figure~\ref{fig:app_soft_token_attention}, the mean budgets are $6.5\%$, $3.3\%$, and $1.0\%$, respectively. The correct token receives a $98\%$ larger budget than the wrong-domain token (bootstrap $95\%$ interval: $92\%$--$104\%$), showing semantic selectivity rather than a position-only effect.

\subsection{Robustness to Source--Target Shift}
\label{app:ood}

The frozen \texttt{sentence-transformers/all-MiniLM-L6-v2} encoder~\citep{reimers2019sentencebert}, which is independent of every compared personalization method, embeds an equal-sized held-out set of interaction inputs from each domain. Let $\mu_d$ be the $\ell_2$-normalized mean embedding. Target--source proximity and shift are
\begin{equation}
\begin{aligned}
s(d_t,\mathcal D_s)
&=
\max_{d_s\in\mathcal D_s}
\cos(\mu_{d_t},\mu_{d_s}),
\\
\operatorname{shift}(d_t)
&=1-s(d_t,\mathcal D_s).
\end{aligned}
\label{eq:app_domain_shift}
\end{equation}
Equal sample sizes prevent domain frequency from changing the estimator's variance. The external encoder and held-out interactions are not used by the domain graph or retrieval module.

The most shifted evaluated targets are LaMP-QA Philosophy ($s=0.156$) and HiCUPID Food ($s=0.302$). Relative to the strongest competing method in each setting, \method improves WR/CCS by $15.4\%/13.9\%$ on Philosophy and $9.6\%/21.6\%$ on Food. Across the five HiCUPID held-out domains, the mean cell-wise relative WR gain is $15.5\%$, compared with $4.1\%$ on the source-domain average. The larger improvement under the evaluated shifts supports the intended transfer behavior, but these observations are not a guarantee for arbitrary domains outside the support of the source data.

\subsection{Latency Breakdown}
\label{app:latency}

Table~\ref{tab:app_latency} reports matched end-to-end latency, including support-statistic computation and all inner-loop updates for the few-shot systems.

\begin{table}[H]
\centering
\small
\setlength{\tabcolsep}{3.0pt}
\begin{tabular}{lcc}
\toprule
Variant & Latency (s) & $\Delta$ vs. vanilla (s) \\
\midrule
Zero-shot generation & 2.45 & -- \\
Vanilla Meta-LoRA, 5-shot & 3.20 & 0.00 \\
Full method, 5-shot & 3.42 & 0.22 \\
\bottomrule
\end{tabular}
\caption{Mean end-to-end latency per response under identical hardware, batch size, sequence length, and decoding settings.}
\label{tab:app_latency}
\end{table}

The full method adds $0.22$ seconds over vanilla Meta-LoRA in this setting and therefore introduces no substantial deployment-latency increase under the matched configuration. Absolute latency remains hardware- and decoding-dependent.

\subsection{Evaluator Agreement Results}
\label{app:evaluator_agreement}

Table~\ref{tab:app_evaluator_agreement} compares the fixed Qwen3-based scores against two method-independent judgment sources on the same blinded samples.

\begin{table}[H]
\centering
\small
\setlength{\tabcolsep}{3.2pt}
\begin{tabular}{lccc}
\toprule
Benchmark & Samples & Human $r$ & GPT-4o $r$ \\
\midrule
HiCUPID & 100 & 0.917 & 0.942 \\
LaMP-QA & 100 & 0.868 & 0.955 \\
\bottomrule
\end{tabular}
\caption{Agreement of the Qwen3-based metrics with method-blinded human and independent GPT-4o judgments.}
\label{tab:app_evaluator_agreement}
\end{table}

All four Pearson correlations are high, indicating that the reported metrics track human-perceived personalization quality and are robust to evaluator choice. This evidence supports the validity of the measurement protocol; it does not imply that automatic evaluation can replace human assessment in every deployment setting.

\subsection{Hyperparameter Sensitivity Protocol}
\label{app:sensitivity}

Sensitivity is evaluated without target-domain tuning. Tables~\ref{tab:app_variance_sensitivity} and~\ref{tab:app_entropy_sensitivity} report the source-validation sweeps used to select the entropy coefficient and the initialization of the bounded, outer-loop-learned group variances. The low--high--low initialization is not a manually fixed final schedule: the three values initialize trainable log-variances, which are subsequently updated only on source episodes. Moderate perturbations preserve a clear advantage, whereas a uniform or inverted initialization is substantially worse. Likewise, both removing entropy modulation and over-scaling it reduce performance, with $\alpha=1$ providing the best joint WR/CCS result. The absolute source-validation scores are not held-out target-test estimates and are reported only to document selection and sensitivity.

\begin{table}[H]
\centering
\small
\setlength{\tabcolsep}{4.0pt}
\begin{tabular}{lcc}
\toprule
Variance initialization & WR $\uparrow$ & CCS $\uparrow$ \\
\midrule
$0.05/0.15/0.08$ (selected) & \textbf{72.9} & \textbf{1.58} \\
$0.08/0.12/0.10$ & 71.7 & 1.46 \\
$0.10/0.10/0.10$ & 68.4 & 1.40 \\
$0.15/0.05/0.12$ & 62.3 & 1.19 \\
Strongest baseline & 63.1 & 1.27 \\
\bottomrule
\end{tabular}
\caption{Source-validation sensitivity to early/middle/late variance initialization. The variances remain trainable on source episodes.}
\label{tab:app_variance_sensitivity}
\end{table}

\begin{table}[H]
\centering
\small
\setlength{\tabcolsep}{5.0pt}
\begin{tabular}{lcc}
\toprule
Entropy coefficient & WR $\uparrow$ & CCS $\uparrow$ \\
\midrule
$\alpha=0$ & 69.6 & 1.41 \\
$\alpha=0.5$ & 70.8 & 1.43 \\
$\alpha=1$ (selected) & \textbf{72.9} & \textbf{1.58} \\
$\alpha=2$ & 70.4 & 1.39 \\
Strongest baseline & 63.1 & 1.27 \\
\bottomrule
\end{tabular}
\caption{Source-validation sensitivity to predictive-entropy modulation.}
\label{tab:app_entropy_sensitivity}
\end{table}

Retrieval additionally evaluates candidate caps $k\in\{1,3,5,8\}$ and source-selected edge thresholds around the default. Topology alignment varies $\lambda_{\mathrm{attn}}$ and the number of soft tokens while keeping the total source training budget fixed. We report WR and CCS together with the number of retained neighbors and learned variance values, because performance alone cannot distinguish a robust region from a degenerate sparse graph.

\begin{figure}[t]
\centering
\includegraphics[width=\columnwidth]{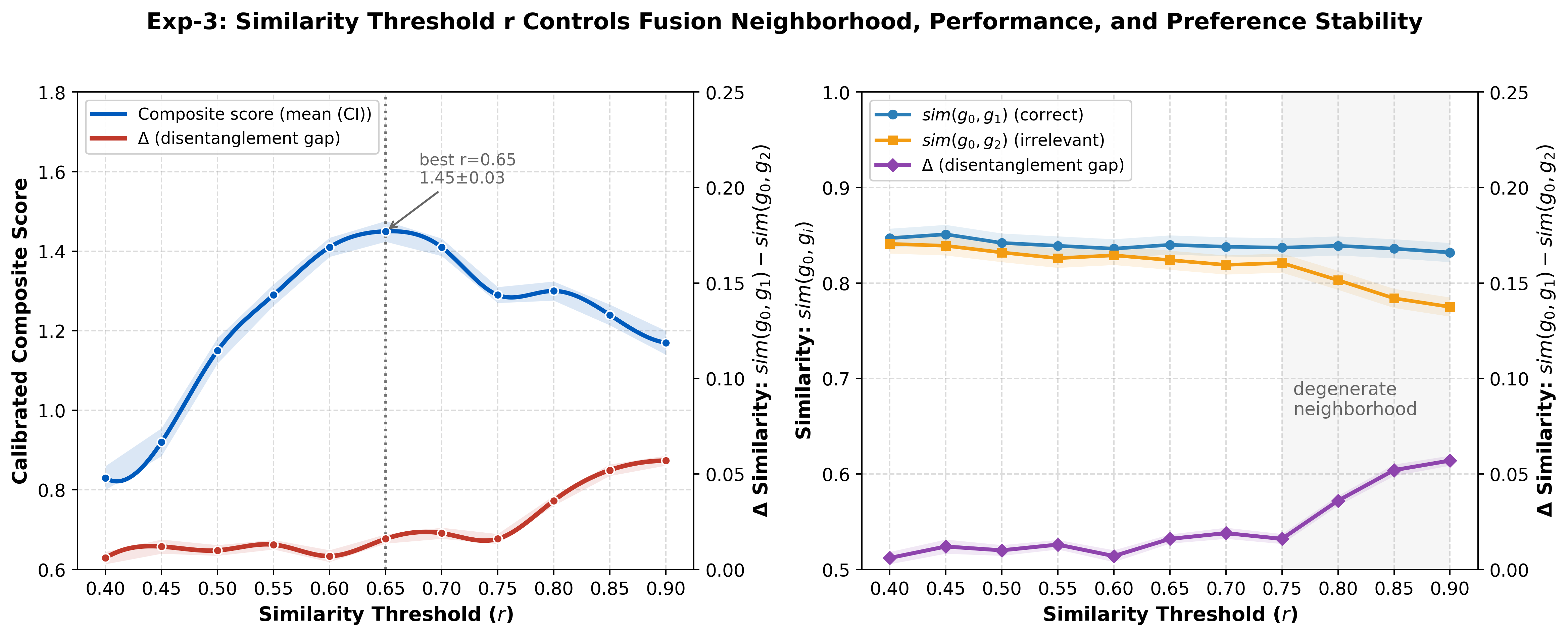}
\caption{Source-validation sensitivity to the graph edge threshold. The left panel reports CCS; the right panel tracks the retained-neighborhood behavior and resistance to irrelevant-domain injection.}
\label{fig:app_threshold_sensitivity}
\end{figure}

The graph-threshold sweep in Figure~\ref{fig:app_threshold_sensitivity} isolates the retrieval precision--coverage trade-off. CCS peaks at the source-selected threshold $0.65$ ($1.45\pm0.03$), where an average of $3.2$ source neighbors is retained. Lower thresholds admit unreliable relations; thresholds above $0.8$ frequently leave a singleton or empty transferable neighborhood. Performance is stable in the $0.60$--$0.70$ region, indicating that the reported result is not tied to a knife-edge choice.

\Needspace{18\baselineskip}
\subsection{Case Study}
\label{app:case}

Table~\ref{tab:app_case} contrasts a meta-learning baseline with the complete framework on one held-out Food-domain query; the preference cues are reconstructed from history and shown only for interpretation.

\begin{table}[H]
\centering
\small
\setlength{\tabcolsep}{3.0pt}
\renewcommand{\arraystretch}{1.08}
\begin{tabular}{@{}p{0.20\columnwidth}p{0.74\columnwidth}@{}}
\toprule
Item & Content \\
\midrule
Query & \emph{How to relax after a busy day?} \\
History-supported cues & Bubble-tea preference; interest in history and urban planning; detail orientation. \\
MAML & ``Reading about urban design could be a relaxing hobby.'' \\
PROPER & ``Reading about urban design can be a structured, relaxing hobby.'' \\
Ours & ``Walk through a historic neighborhood, pair it with a history read, and end with your favorite bubble tea.'' \\
\bottomrule
\end{tabular}
\caption{Illustrative Food-domain case. Preference cues are shown only for analysis and are not supplied as oracle input.}
\label{tab:app_case}
\end{table}

The two baselines recover the broad urban-design topic but omit the jointly supported history, detail-orientation, and bubble-tea cues. The complete method coordinates those cues in one domain-appropriate suggestion. This is qualitative evidence only; aggregate conclusions are based on the quantitative test sets.

\subsection{Result Scope and Traceability}
\label{app:full_results}

The main-paper tables are the authoritative report of target-domain predictive performance. Appendix H adds the full Qwen3 cross-backbone table and analyses of components in the submitted method: nested-shot efficiency, source-training stability, topology and attention diagnostics, source--target shift, latency, evaluator agreement, source-validation sensitivity, and the qualitative case. The code archive fixes the final backbone identifiers, domain splits, seeds, metric formulas, and component configurations, providing one traceable specification for every reported analysis.

\section{Reproducibility, Limitations, and Responsible Use}
\label{app:responsibility}

\subsection{Reproducibility Package}
\label{app:artifacts}

The accompanying materials contain this technical appendix and a separate code archive. Table~\ref{tab:artifact_manifest} maps its contents to the paper. The archive uses relative paths and includes an English README plus a short evaluation guide with executable entry points for preprocessing, source-only prior construction, episodic training, target adaptation, evaluation, statistical testing, and core-equation verification.

\begin{table*}[t]
\centering
\small
\setlength{\tabcolsep}{4.0pt}
\renewcommand{\arraystretch}{1.06}
\begin{tabular}{p{0.27\textwidth}p{0.39\textwidth}p{0.27\textwidth}}
\toprule
Archive path & Contents & Paper reference \\
\midrule
\texttt{configs/} & Final proposed-method and baseline configurations, development search spaces, seeds, and model identifiers & Appendix~\ref{app:baselines} and Tables~\ref{tab:baseline_hyperparameters}--\ref{tab:hyperparameter_search} \\
\texttt{scripts/prepare\_data.py} & Deterministic normalization, domain-disjoint partitioning, and split-manifest generation for legally obtained benchmark files & Appendix~\ref{app:datasets} \\
\texttt{src/.../preference/} & History perturbations, source-only rewards, GRPO objective, and textual prompt generation & Appendix~\ref{app:summarizer} \\
\texttt{src/.../domain/} & Geometry-compatible features, reliability graphs, Sinkhorn barycenter, leave-one-domain-out composition, and topology-preserving projection & Appendix~\ref{app:domain} \\
\texttt{src/.../adaptation/} & PAC-Bayes utilities, entropy and Fisher statistics, group variances, constrained inner loop, and episodic meta-learning & Eqs.~\eqref{eq:app_gamma}--\eqref{eq:app_adapt} \\
\texttt{src/.../pipeline/} & Source training and unified zero-/few-shot inference with explicit information boundaries & Algorithms~\ref{alg:app_source_training}--\ref{alg:app_target_inference} \\
\texttt{src/.../evaluation/} & WR, CCS, PG, blinded ordering, paired tests, Holm correction, and bootstrap intervals & Appendix~\ref{app:metrics}--\ref{app:statistics} \\
\texttt{prompts/} and \texttt{results/} & Frozen evaluator/summarizer prompts and structured Qwen3, sensitivity, latency, agreement, and OOD results & Appendices~\ref{app:qwen3_results}--\ref{app:sensitivity} \\
\texttt{configs/baselines/} & Method-specific settings preserving data, backbone, decoding, and support-budget comparability & Appendix~\ref{app:baselines} \\
\texttt{tests/} and \texttt{examples/} & Unit tests and synthetic inputs for a CPU smoke test & Appendices~\ref{app:pac}--\ref{app:algorithms} \\
\texttt{environment/} & Locked dependencies and the reported hardware/software inventory & Appendix~\ref{app:hyperparameters} \\
\texttt{ARTIFACT\_...\_GUIDE.md} & Short structural, CPU, and full-reproduction audit paths plus a claim-to-code map & Appendix~\ref{app:checklist_map} \\
\texttt{DATA\_CARD.md}, \texttt{MODEL\_CARD.md} & Data boundaries, intended use, limitations, privacy, safety, and non-redistribution scope & Appendices~\ref{app:datasets}, \ref{app:limitations}--\ref{app:safety} \\
\bottomrule
\end{tabular}
\caption{Manifest of the accompanying code archive. The listed files are contained in the released artifact.}
\label{tab:artifact_manifest}
\end{table*}

The archive additionally contains:
\begin{itemize}
\item deterministic utilities that create source/target and nested-support manifests from public benchmark files;
\item the exact environment, library versions, hardware record, final configuration, and complete development search space;
\item source-training, target-adaptation, and evaluation entry points;
\item evaluator prompts, response-order randomization code, and tie policy;
\item the ten declared random seeds and implementations of paired confidence intervals and corrected tests;
\item a synthetic smoke test that verifies the central equations without redistributing benchmark text or model weights.
\end{itemize}

Original code is licensed under Apache-2.0. Dataset files, pretrained checkpoints, and third-party baselines are not relicensed or redistributed; each remains governed by its public release terms. The code archive uses relative paths and excludes local filesystem paths, private repository references, and machine-specific metadata. The included tests and smoke example run independently of benchmark text and pretrained model downloads.

\subsection{Checklist-to-Artifact Cross-Reference}
\label{app:checklist_map}

The conceptual method and complete inference path requested by the
reproducibility checklist are provided in Appendices~\ref{app:pac}
through~\ref{app:algorithms}.  The status and scope of the theoretical
statements are delineated in Appendix~\ref{app:pac}; Proposition~1 is a
standard randomized-predictor PAC-Bayes result, whereas the entropy-, Fisher-,
and depth-calibrated deterministic objective is an empirically evaluated
surrogate rather than an additional guarantee.  Dataset motivation and splits,
baseline settings, metric definitions, development ranges and selection
criterion, randomness controls, computing infrastructure, run counts,
variation, and statistical tests are recorded in
Appendix~\ref{app:experiments}.  The archive paths supporting preprocessing,
training, evaluation, analysis, licensing, and code-to-paper comments are
listed in Table~\ref{tab:artifact_manifest}.

\subsection{Limitations}
\label{app:limitations}

The PAC-Bayes statement concerns a randomized predictor under i.i.d.\ bounded-loss assumptions; deployment uses a deterministic surrogate. Predictive entropy can be miscalibrated, and a squared-gradient Fisher proxy is only a local sensitivity estimate. The domain prior can fail when descriptions are ambiguous or every source domain is irrelevant. A textual preference prompt may omit subtle or changing preferences, while continuous domain tokens are not directly human-readable. With no pre-target history, the method falls back to a neutral textual prompt and domain conditioning, so it cannot infer genuinely user-specific tendencies until user evidence becomes available. These limitations constrain the claims and motivate calibration, abstention, and longitudinal evaluation.

\subsection{Privacy and Data Governance}
\label{app:privacy}

The experiments use anonymized benchmark signals. The method does not require cross-user graph propagation in its current form. Deployment should be opt-in and should provide a way to inspect, edit, or delete the textual user prompt and associated target adapters. Raw histories should be retained only as required by the application, and cached prompts or adapters should be access-controlled and scoped to one user. The method does not constitute a formal privacy guarantee; differential privacy, membership-inference resistance, and secure deletion remain separate engineering and research requirements.

\subsection{Safety and Misuse}
\label{app:safety}

Personalization may amplify an incorrect inferred preference, create filter bubbles, or overfit sensitive attributes. The generation policy must continue to obey task-level safety constraints even when a user prompt or domain token suggests otherwise. Applications should distinguish user-editable preferences from protected or sensitive inferences, use conservative defaults when evidence is weak, and permit non-personalized fallback. Human agreement with an evaluator supports metric validity but does not establish safety in high-stakes domains.

\end{document}